\documentclass[sigconf]{acmart}
\AtBeginDocument{%
  }

\setcopyright{none}
\copyrightyear{2025}
\acmYear{2025}
\acmDOI{XXXXXXX.XXXXXXX}
\thanks{${*}$Equal contribution. $\dagger$Corresponding author: Xin Liu (email: xin.liu@lut.fi)}
\acmConference[ACM MM]{the 33rd ACM International Conference on Multimedia}{October 27--31, 2025}{Dublin, Ireland}
\acmISBN{978-1-4503-XXXX-X/2018/06}

\usepackage{multirow} 
\usepackage{algorithm,algpseudocode}
\usepackage{comment}
\usepackage{xcolor}
\usepackage{hyperref}
\usepackage[export]{adjustbox}

\definecolor{nfbl}{HTML}{97ad66}
\definecolor{video}{HTML}{acc3f2}
\definecolor{audio}{HTML}{e67086}
\definecolor{trans}{HTML}{7290b8}
\definecolor{correct}{HTML}{41BF87}

\begin{document}

\title{DEEMO: De-identity Multimodal Emotion Recognition and Reasoning}


\author{Deng Li$^{*}$}
\affiliation{%
  \institution{Lappeenranta-Lahti University of Technology LUT}
  \city{Lappeenranta}
  \country{Finland}}
\email{deng.li@lut.fi}

\author{Bohao Xing$^{*}$}
\affiliation{%
  \institution{Lappeenranta-Lahti University of
Technology LUT}
  \city{Lappeenranta}
  \country{Finland}}
\email{bohao.xing@lut.fi}

\author{Xin Liu$^\dagger$}
\affiliation{%
  \institution{Lappeenranta-Lahti University of Technology LUT}
  \city{Lappeenranta}
  \country{Finland}}
\email{xin.liu@lut.fi}

\author{Baiqiang Xia}
\affiliation{%
  \institution{Silo AI}
  \city{Helsinki}
  \country{Finland}}
\email{baiqiang.xia@silo.ai}

\author{Bihan Wen}
\affiliation{%
  \institution{Nanyang Technological University}
  \city{Singapore}
  \country{Singapore}}
\email{bihan.wen@ntu.edu.sg}

\author{Heikki Kälviäinen}
\affiliation{%
  \institution{Lappeenranta-Lahti University of Technology LUT}
  \city{Lappeenranta}
  \country{Finland}\\
  \institution{Brno University of Technology}
  \city{Brno}
  \country{Czech Republic}}
\email{heikki.kalviainen@lut.fi}

\renewcommand{\shortauthors}{Deng Li et al.}

\begin{abstract}

Emotion understanding is a critical yet challenging task. Most existing approaches rely heavily on identity-sensitive information, such as facial expressions and speech, which raises concerns about personal privacy. To address this, we introduce the \textbf{De}-identity Multimodal \textbf{Emo}tion Recognition and Reasoning (\textbf{\textit{DEEMO}}), a novel task designed to enable emotion understanding using de-identified video and audio inputs. The \textit{DEEMO} dataset consists of two subsets: \textbf{\textit{DEEMO-NFBL}}, which includes rich annotations of \textbf{N}on-\textbf{F}acial \textbf{B}ody \textbf{L}anguage (NFBL), and \textbf{\textit{DEEMO-MER}}, an instruction dataset for \textbf{M}ultimodal \textbf{E}motion \textbf{R}ecognition and \textbf{R}easoning using identity-free cues. This design supports emotion understanding without compromising identity privacy. In addition, we propose DEEMO-LLaMA, a Multimodal Large Language Model (MLLM) that integrates de-identified audio, video, and textual information to enhance both emotion recognition and reasoning. Extensive experiments show that DEEMO-LLaMA achieves state-of-the-art performance on both tasks, outperforming existing MLLMs by a significant margin, achieving 74.49\% accuracy and 74.45\% F1-score in de-identity emotion recognition, and 6.20 clue overlap and 7.66 label overlap in de-identity emotion reasoning. Our work contributes to ethical AI by advancing privacy-preserving emotion understanding and promoting responsible affective computing.

\end{abstract}





\maketitle

\section{Introduction}
\begin{figure}[t]
    \centering
    \footnotesize
    \includegraphics[width=0.9\linewidth]{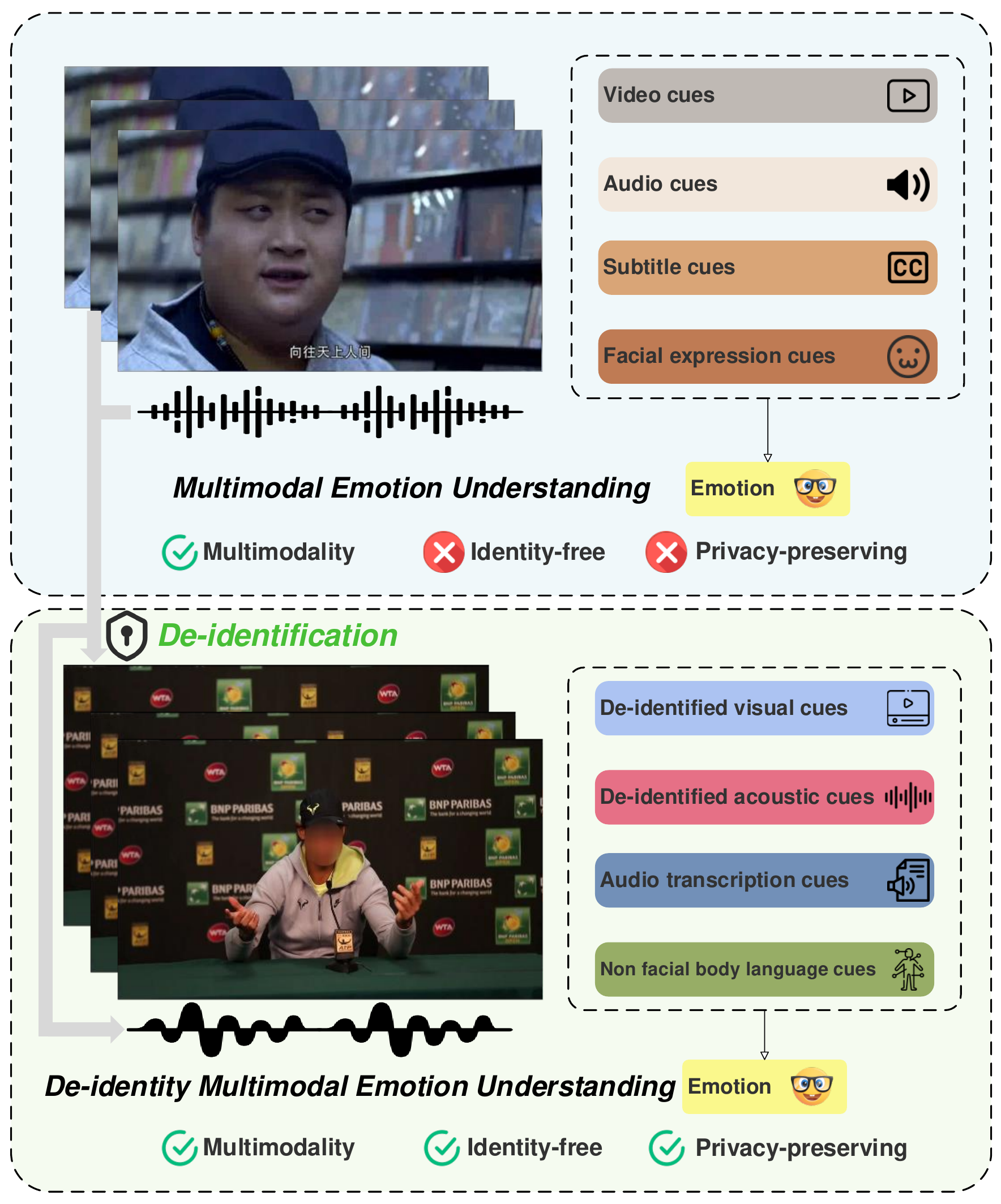}
    \vspace{-1em}
    \caption{Previous multimodal emotion understanding tends to rely on identity-sensitive information (e.g., facial expression). In contrast, \textit{DEEMO} leverages de-identified video and audio, identity-free audio transcriptions, and non-facial body language (NFBL) cues to enable privacy-preserving emotion recognition and reasoning.}
    \vspace{-3em}
    \label{fig:motivation}
\end{figure}
Emotion understanding is one of the most fundamental yet challenging tasks~\cite{koelstra2011deap, hakak2017emotion}. In recent decades, emotion understanding has attracted a lot of attention from the research community~\cite{nandwani2021review, li2020deep, el2011survey, ezzameli2023emotion, rahdari2019multimodal, xing2024emo}. However, most existing approaches rely on identity-sensitive information, such as facial expressions~\cite{wei2024learning} and speech~\cite{rathi2024analyzing}, as summarized in Table~\ref{tab:datasets}. With growing societal concerns over privacy~\cite{meden2021privacy} and increasing user awareness around personal data sharing\cite{regulation2018general}, it has become increasingly important to develop emotion understanding methods that do not depend on identity-sensitive information, as shown in Figure~\ref{fig:motivation}.

To explore emotion understanding beyond identity-related features, we propose novel tasks called \textbf{De}-identity Multimodal \textbf{Emo}tion Recognition and Reasoning and introduce the corresponding dataset, \textbf{\textit{DEEMO}}. Our goal is to advance emotion understanding under privacy constraints. To this end, the \textit{DEEMO} dataset explicitly promotes privacy preservation by excluding identity-sensitive modalities such as facial image/video and raw speech. Instead, it leverages de-identified video and audio to enable emotion recognition and reasoning in a privacy-preserving manner. Specifically, we collect post-match press interview videos, where professional athletes from various sports (e.g., tennis, football, and basketball) respond to multiple rounds of questions following intense competitions. To enrich the dataset with high-quality recognition and reasoning annotations, we adopt a semi-automatic pipeline that combines advanced Large Language Models (LLMs) with human review. This approach allows \textit{DEEMO} to provide both standard emotion recognition labels and detailed reasoning based on de-identified visual cues, de-identified acoustic cues, audio transcription cues, and Non-Facial Body Language (NFBL) cues, as shown in Figure~\ref{fig:motivation}. 
The \textit{DEEMO} consists of two subsets: \textit{DEEMO-NFBL}, which contains 24,722 NFBL annotations of 37 classes, and \textit{DEEMO-MER}, which includes 2,060 annotated videos with both emotion recognition labels and reasoning instructions via LLM-human collaborative annotation.
\begin{table}[t]
    \centering
    \footnotesize
    \setlength\tabcolsep{1pt}
    \renewcommand{\arraystretch}{1.2}
    \caption{Comparison of the different emotion datasets. V, A, and T denote video, audio, and text, respectively.}
    \vspace{-1em}
    \begin{tabular}{llrrrr}
    \toprule    
    \multirow{3}{*}{\textbf{Dataset}}&\multirow{3}{*}{\textbf{Modality}}&\multicolumn{2}{c}{\textbf{Tasks}}&\multicolumn{2}{c}{\textbf{De-identity}}\\ \cmidrule(lr){3-4} \cmidrule(lr){5-6} 
    &&\textbf{Emotion}&\textbf{Emotion}&\textbf{NFBL}&\textbf{Identity}\\
    &&\textbf{Recognition?}&\textbf{Reasoning?}&\textbf{Annota.}?&\textbf{free?}  \\
    \hline
    HUMAINE~\cite{douglas2007humaine}&V+A+T&\checkmark&$\times$&\checkmark&$\times$\\
    VAM~\cite{grimm2008vera}&V+A&\checkmark&$\times$&$\times$&$\times$\\
    IEMOCAP~\cite{busso2008iemocap}&V+A+T&\checkmark&$\times$&$\times$&$\times$\\
    Youtube~\cite{morency2011towards}&V+A+T&\checkmark&$\times$&$\times$&$\times$\\
    AFEW~\cite{dhall2012collecting}&V+A&\checkmark&$\times$&$\times$&$\times$\\
    AM-FED~\cite{mcduff2013affectiva}&V&\checkmark&$\times$&$\times$&$\times$\\
    AFEW-VA~\cite{dhall2015video}&V+A&\checkmark&$\times$&$\times$&$\times$\\
    LIRIS-ACCEDE~\cite{baveye2015liris}&V+T&\checkmark&$\times$&\checkmark&$\times$\\
    EMILYA~\cite{fourati2014emilya}&V+T&\checkmark&$\times$&\checkmark&$\times$\\
    SEWA~\cite{kossaifi2019sewa}&V+A&\checkmark&$\times$&$\times$&$\times$\\
    CMU-MOSEI~\cite{zadeh2018multimodal}&V+A+T&\checkmark&$\times$&$\times$&$\times$\\
    iMiGUE~\cite{liu2021imigue}&V+T&\checkmark&$\times$&\checkmark&\checkmark\\
    MERR~\cite{cheng2025emotion}&V+A+T&\checkmark&$\checkmark$&$\times$&$\times$\\
    EMER~\cite{lian2024affectgpt}&V+A+T&\checkmark&$\checkmark$&$\times$&$\times$\\
    \textbf{DEEMO (Ours)}&V+A+T&\checkmark&$\checkmark$&\checkmark&\checkmark\\
    \bottomrule
    \end{tabular}
    \label{tab:datasets}
    \vspace{-1em}
\end{table}

To facilitate this task, we further propose DEEMO-LLaMA, a framework that integrates multimodal de-identified cues for both emotion recognition and reasoning. Through extensive experiments, we demonstrate that our approach outperforms existing baselines under the de-identified setting.
We hope that \textit{DEEMO} can serve as a foundation for future research in privacy-preserving affective computing and responsible AI. Our main contributions are summarized as follows:
\vspace{-0.5em}
\begin{itemize}
\item We propose new tasks: de-identity multimodal emotion recognition and reasoning. To support this task, we construct the corresponding \textit{DEEMO} dataset. It consists of two subsets: 1) \textit{DEEMO-NFBL}, which includes 37 classes of NFBL with a total of 24,722 annotations; 2) \textit{DEEMO-MER}, which contains 2,060 videos annotated with both emotion recognition and reasoning instructions. Unlike previous datasets that rely on identity-sensitive modalities, \textit{DEEMO} is built entirely from de-identified video, audio, and transcriptions, making it the first dataset specifically designed for privacy-preserving emotion understanding. 
\item We propose DEEMO-LLaMA, a novel multimodal framework for de-identified emotion recognition and reasoning. It integrates three modalities: 1) de-identified video, 2) de-identified audio, and 3) audio transcription. These signals are fused and aligned within a Multimodal Large Language Model (MLLM) architecture, enabling both categorical emotion recognition and open-ended reasoning under privacy constraints.
\item Extensive experiments demonstrate that DEEMO-LLaMA outperforms the existing MLLMs under the de-identity setting. Specifically, it achieves state-of-the-art performance on both emotion recognition (74.49\% accuracy and 74.45\% F1-score) and emotion reasoning (6.20 clue overlap and 7.66 label overlap). 
\end{itemize}

\section{Related Work}

\subsection{Emotion Understanding Datasets}
Various signals have been explored for emotion understanding, including facial expressions~\cite{wei2024learning,canal2022survey,krumhuber2023role} and speech~\cite{rathi2024analyzing,singh2022systematic,hashem2023speech}. 
These biological signals provide valuable cues: facial expressions reflect subtle affective states, while speech encodes emotions through prosody and vocal characteristics. 
However, these signals are inherently identity-sensitive and often raise privacy concerns when deployed in real-world applications. 
In addition, physiological signals such as electrocardiogram (ECG)~\cite{hsu2017automatic}, electroencephalogram (EEG)~\cite{li2022eeg}, and Galvanic Skin Response (GSR)~\cite{liu2016retracted} have also been investigated to capture implicit emotional states. 
However, the acquisition of such signals is often intrusive and impractical in daily interactive scenarios, limiting their applicability. NFBL also plays a crucial role in conveying emotions, particularly in nonverbal communication. Recent efforts, including datasets such as iMiGUE~\cite{liu2021imigue}, SMG~\cite{chen2023smg}, and MA-52~\cite{guo2024benchmarking}, have explored NFBL as an alternative modality for emotion understanding\cite{gao2024identity, li2024enhancing}. 
These works highlight the potential of body language in expressing affective states. However, most existing studies still focus primarily on the recognition of NFBL, while progress on leveraging these cues for higher-level emotion understanding remains relatively limited. Our work builds upon these insights by focusing on identity-free emotional cues, particularly NFBL. \textit{DEEMO} promotes privacy-preserving affective computing by leveraging de-identified video and audio inputs alongside NFBL annotations. This allows us to explore emotion understanding in a way that balances expressiveness with privacy.

\subsection{Multimodal Emotion Understanding}
Previous approaches to emotion understanding have often focused on unimodal signals~\cite{wei2024learning,hashem2023speech} or relied on relatively simple multimodal fusion techniques~\cite{liu2021comparing,middya2022deep}, which struggle to capture the complexity and subtle interplay of emotional cues across modalities. These limitations have hindered the performance and generalization of emotion recognition systems. With the recent advancements in LLMs~\cite{naveed2023comprehensive}, a new direction of research has emerged that leverages the powerful reasoning and representation capabilities of LLMs for emotion understanding. These models enable the integration of rich contextual, semantic, and multimodal information, making it feasible to perform complex affective reasoning. Representative works in this direction include AffectGPT~\cite{lian2024affectgpt}, EMER~\cite{lian2023explainable}, and Emotion-LLAMA~\cite{cheng2025emotion}, which explore how LLMs can be adapted or extended to recognize and interpret emotions across video, audio, and text inputs. These approaches mark a significant shift toward more flexible and scalable emotion understanding frameworks, where LLMs act not only as language processors but also as central reasoning engines for affective cognition.
However, these methods also face privacy concerns, particularly in the visual modality, where they primarily rely on facial information. In contrast, our approach de-identifies multimodal inputs to MLLMs and places greater emphasis on NFBL in the visual modality, providing a more privacy-preserving and ethically aligned solution for emotion understanding.

\section{DEEMO Dataset}
In this section, we present the construction and overall structure of the \textit{DEEMO} dataset, which consists of two primary subsets: 1) \textit{DEEMO-NFBL}, containing a large number of annotated NFBL instances; 2) \textit{DEEMO-MER}, which provides fine-grained annotations for de-identity multimodal emotion recognition and reasoning.

\subsection{\textbf{Data Collection}}
In the \textit{DEEMO} dataset, we use videos of athletes participating in post-game English interviews as the primary data source. This scenario is particularly well-suited for studying emotion understanding for several reasons: 1) The outcome of the game, whether it’s a victory or a loss, serves as a natural trigger for emotions, eliciting either positive or negative states in the interviewed player;  2) The players had no (or very little) time to prepare because the press conference would be held immediately after the game, and athletes needed to respond to the questions rapidly. Unlike acting in movies or series, athletes’ NFBL is natural; 3) Athletes come from diverse cultural and geographic backgrounds, contributing to the dataset's demographic diversity. Based on these factors, we collect 500 post-game interview videos from 6 sports domains such as tennis, football, basketball, boxing, American football, and UFC via YouTube.

\begin{figure}[t]
    \centering
    \tiny
    \setlength\tabcolsep{0.5pt}
    \begin{tabular}{ccc}
        \includegraphics[width=0.3\linewidth]{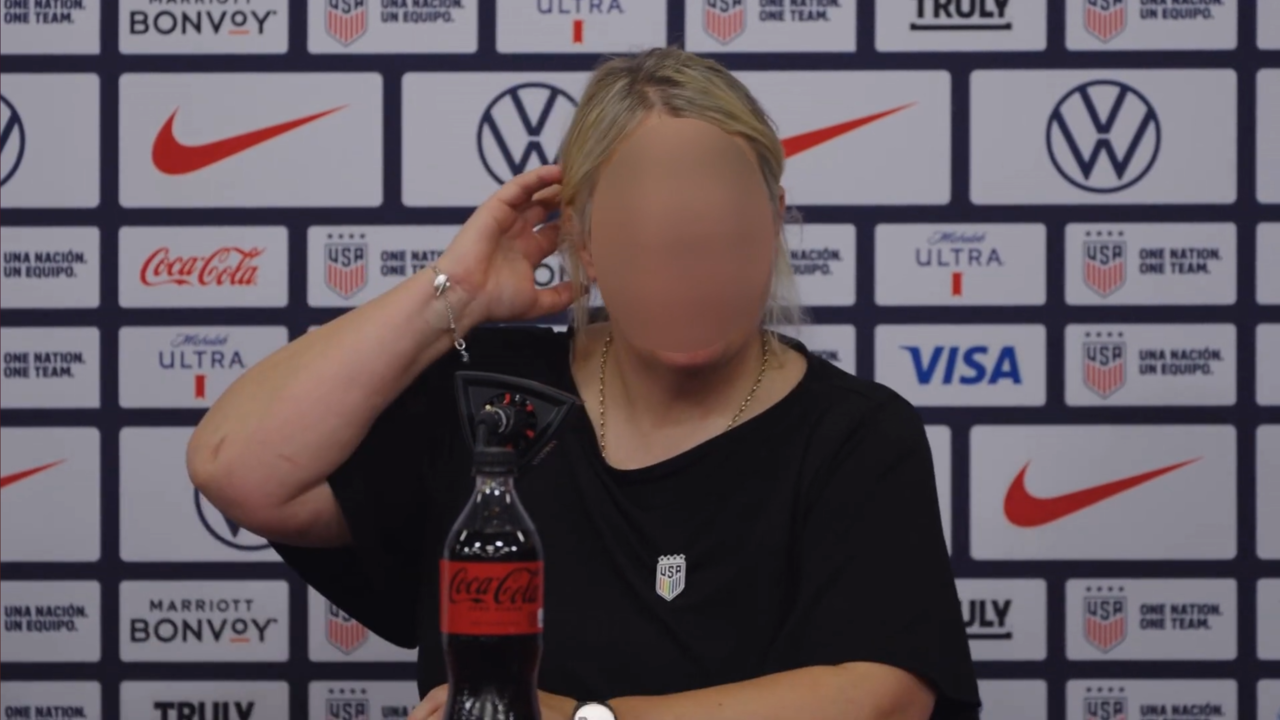}&\includegraphics[width=0.3\linewidth]{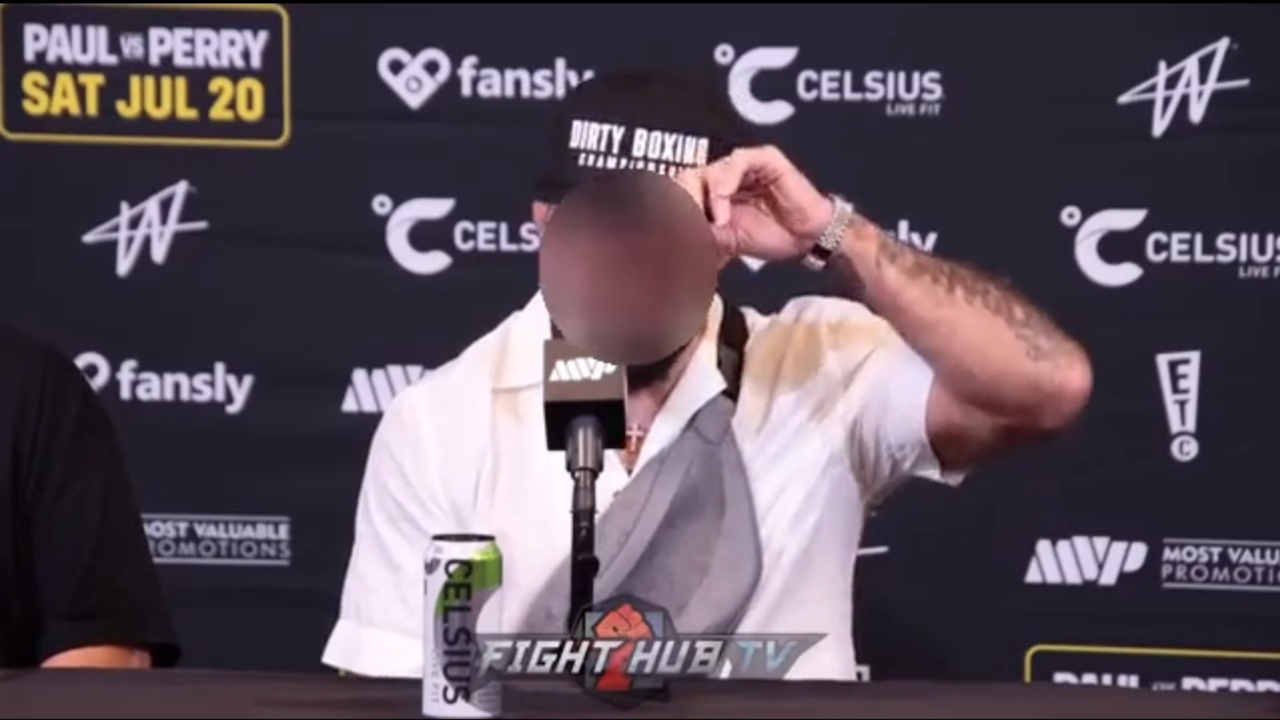}&\includegraphics[width=0.3\linewidth]{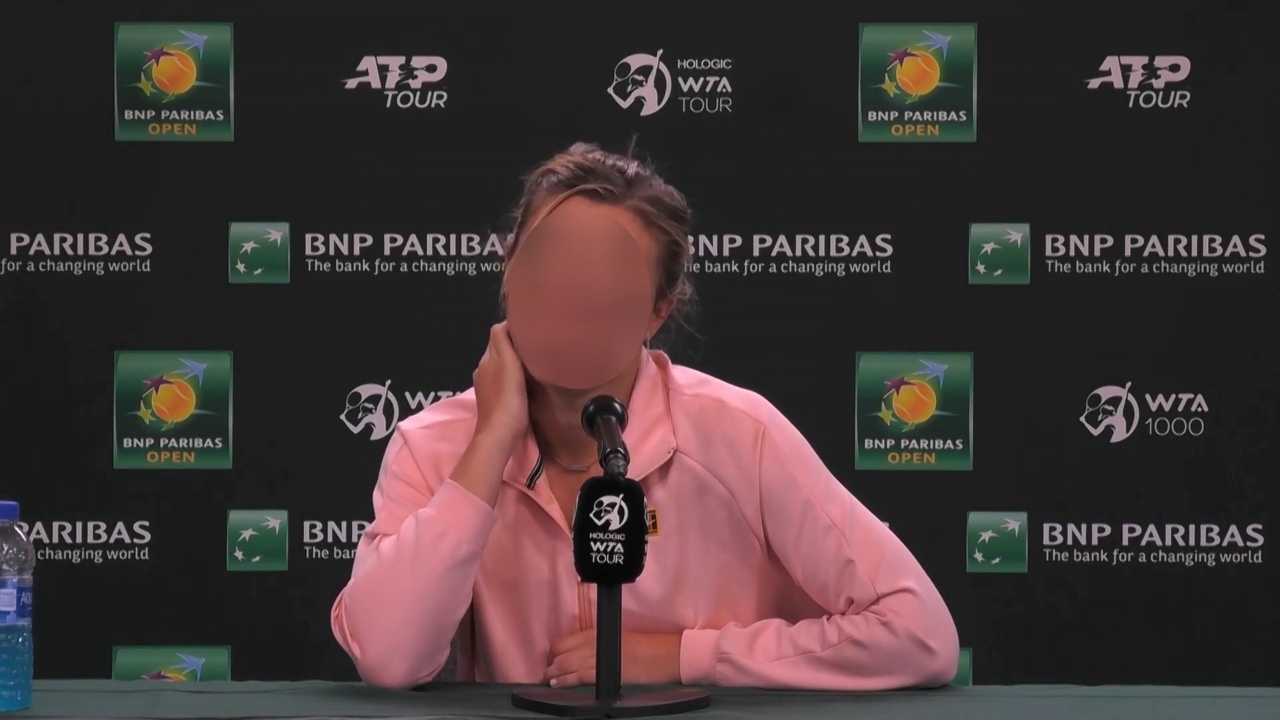}\\
        (a) Touching or scratching head&(b) Touching hat&(c) Touching or scratching neck\\
        \includegraphics[width=0.3\linewidth]{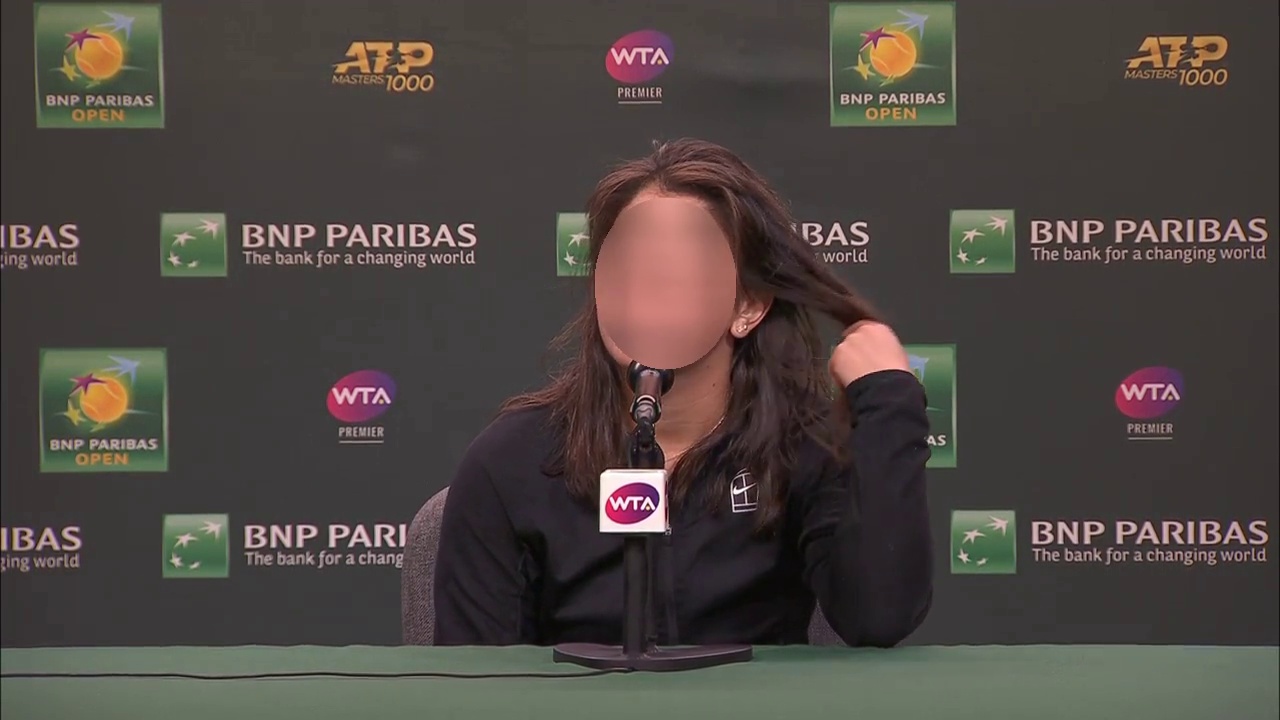}&\includegraphics[width=0.3\linewidth]{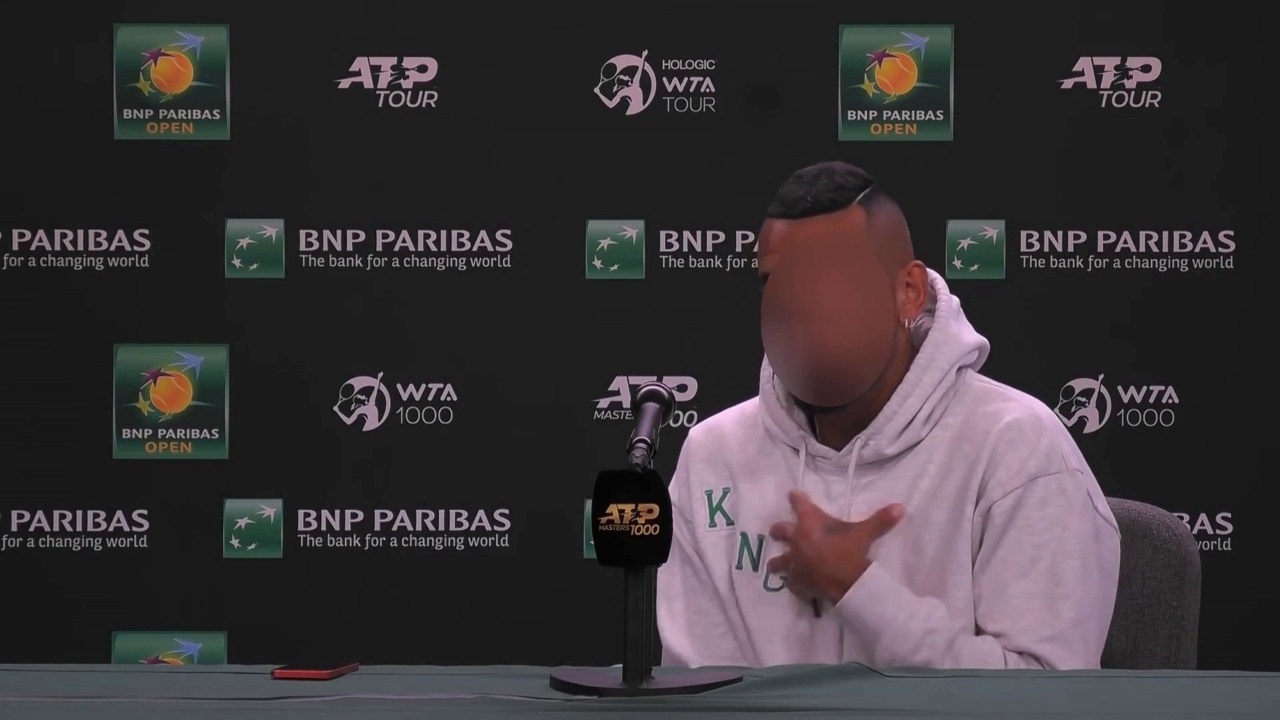}&\includegraphics[width=0.3\linewidth]{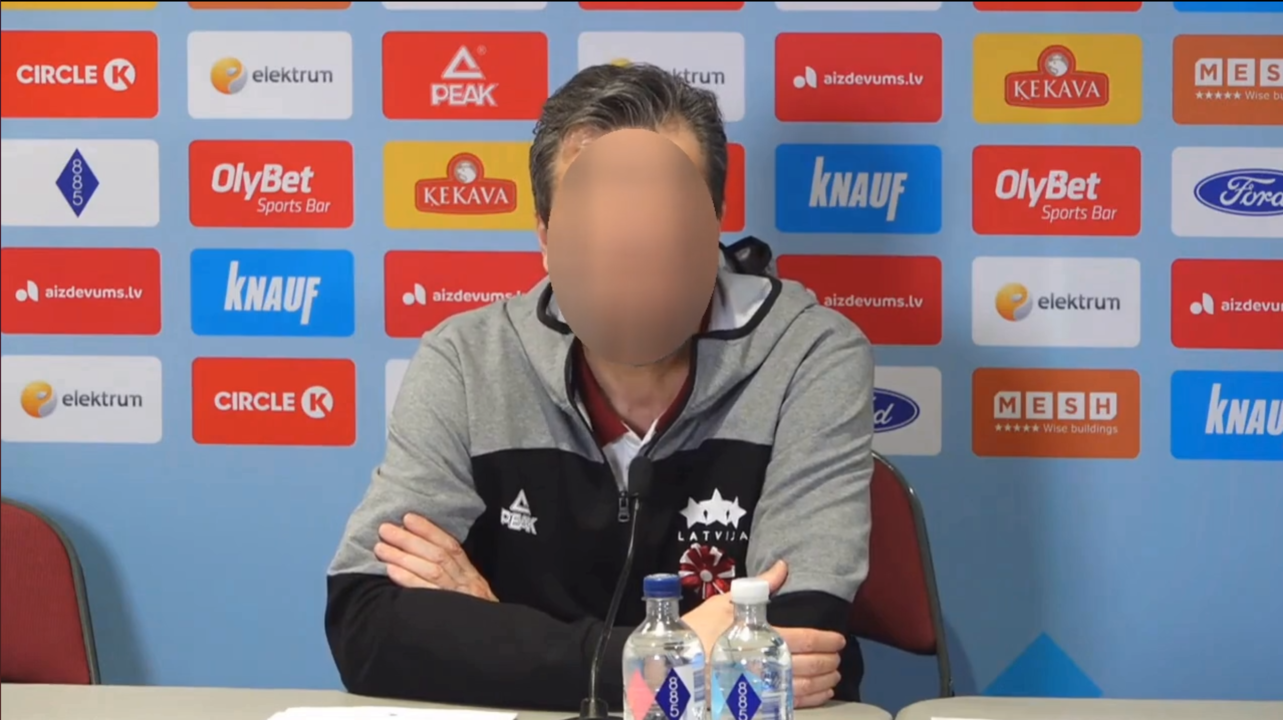}\\
        (d) Playing or adjusting hair&(e) Touching suprasternal notch&(f) Folding arms\\
        \includegraphics[width=0.3\linewidth]{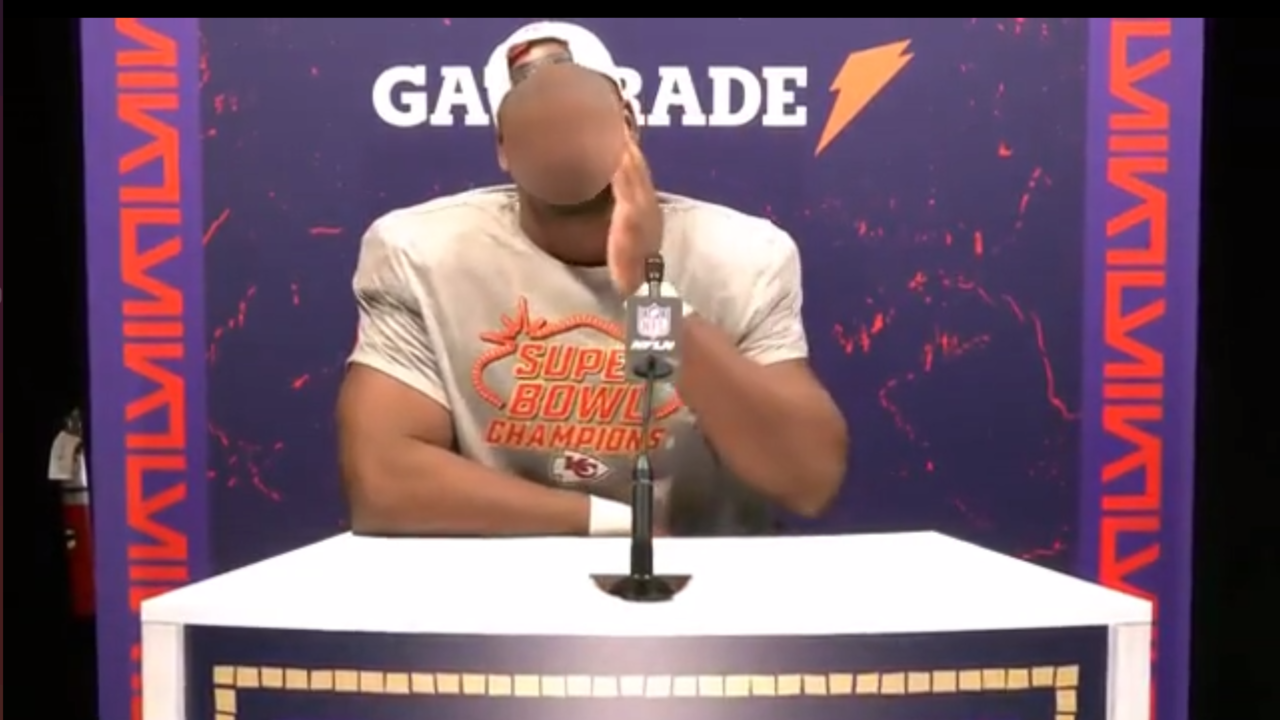}&\includegraphics[width=0.3\linewidth]{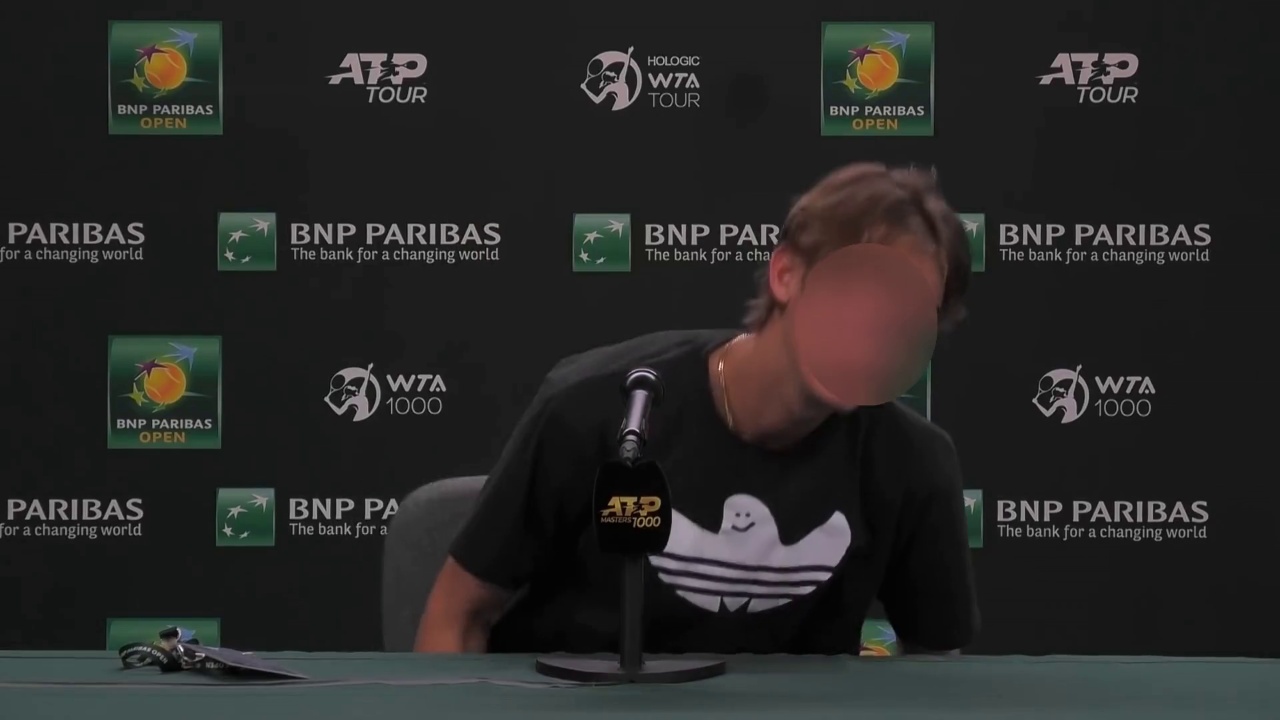}&\includegraphics[width=0.3\linewidth]{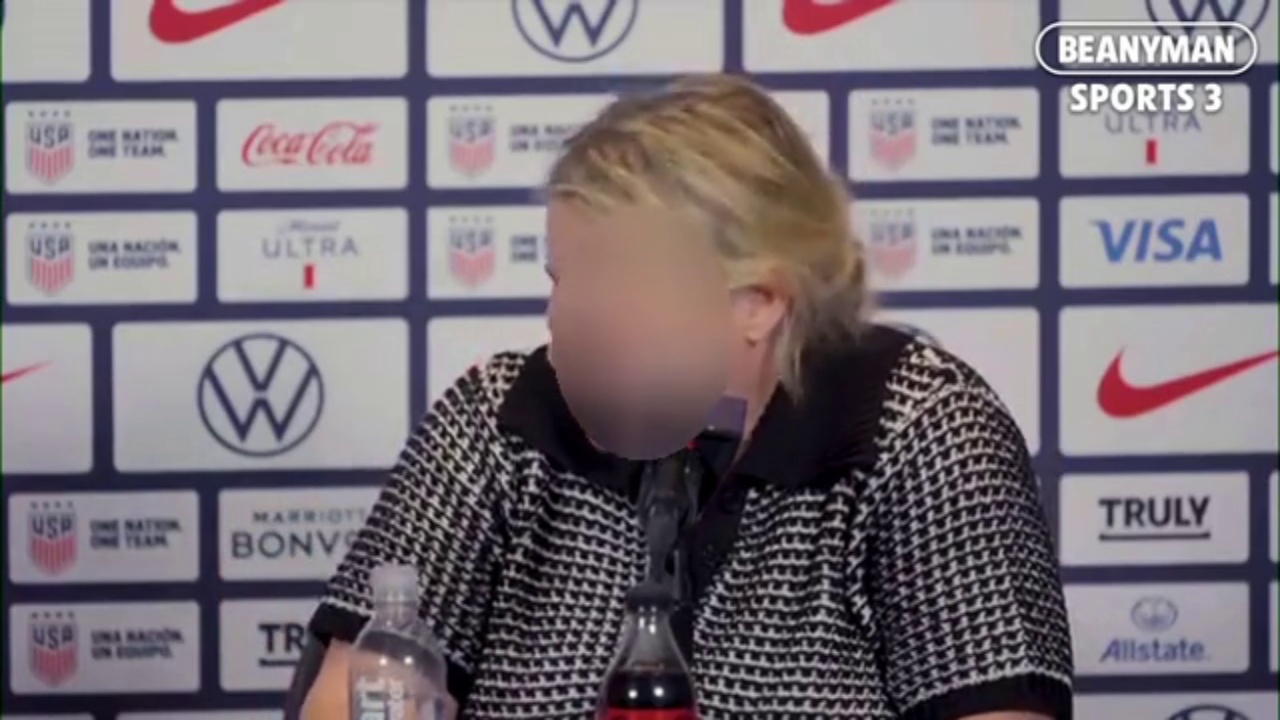}\\
        (g) Scratching facial parts&(h) Moving torso&(i) Turtle neck\\
    \end{tabular}
    \vspace{-1em}
    \caption{Exapmles of \textit{DEEMO-NFBL}.}
    \label{fig:examples}
    \vspace{-2em}
\end{figure}

\subsubsection*{\textbf{Data De-identification}}
The primary objective of the proposed \textit{DEEMO} dataset is to facilitate the study of emotion understanding while preserving privacy. As noted in prior work~\cite{sanderson2004use}, most identity information is embedded in facial and speech signals. To address this, we apply both video and audio de-identification techniques to remove identity information.

For video de-identification, we employ the CenterFace model~\cite{xu2020centerface}, denoted as $m_{\text{face}}$, to locate facial regions in each video frame. Formally, given a video sequence $V = \{v_1, v_2, \dots, v_{N_{\text{sample}}}\}$, where $v_i$ denotes the $i$-th frame, the facial bounding box coordinates $G$ are obtained as $G = m_{\text{face}}(v_i)$. We then apply a Gaussian blur over the detected regions based on $G$ to produce the de-identified video, denoted as $V^{\text{de}}$. Examples of de-identified video frames from the subset \textit{DEEMO-NFBL} are shown in Fig.~\ref{fig:examples}.

Regarding audio de-identification, our goal is to remove speaker identity information while preserving the emotional content of the audio. Recent work on anonymization-based Speech Emotion Recognition (SER) aligns well with this objective. These approaches focus on modifying the speaker’s voice to conceal identity as effectively as possible while preserving both the linguistic content and paralinguistic cues essential for emotion recognition. To this end, we adopt the method proposed in~\cite{tomashenko2024voiceprivacy} for audio de-identification. More specifically, the de-identification process adjusts the spectral envelope of the speech signal using the McAdams coefficient~\cite{mcadams1984spectral} to alter speaker-specific characteristics without significantly distorting emotion-related cues. The resulting de-identified audio is denoted as $A^{\text{de}}$. 

\begin{figure}
    \centering
    \footnotesize
    \setlength\tabcolsep{1pt}
     \includegraphics[clip, trim=6cm 2cm 6cm 1cm,width=\linewidth]{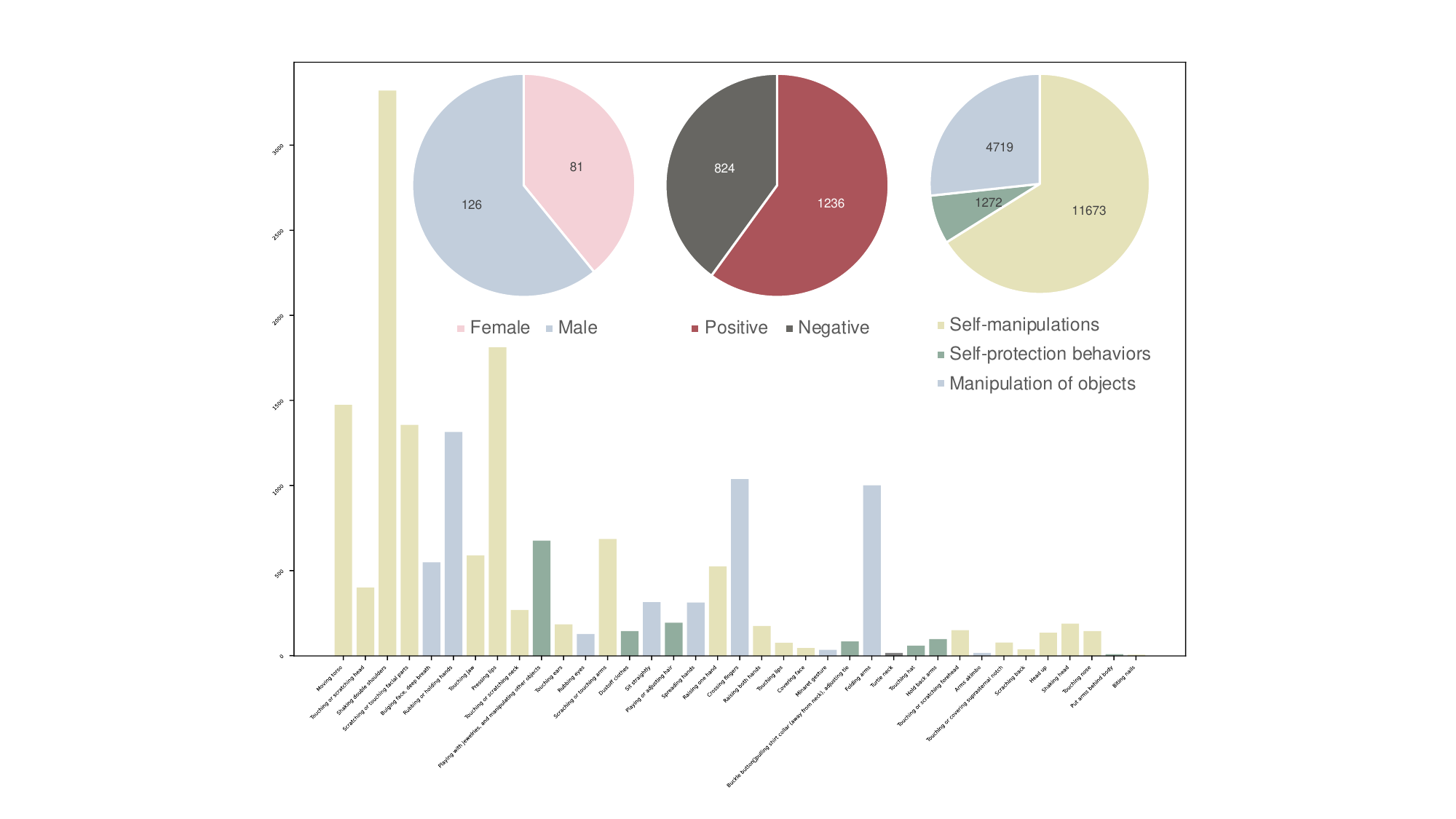}  \\
     \vspace{-1em}
    \caption{Statistical visualization of the \textit{DEEMO} dataset. The bar chart displays the frequency of 37 defined NFBL categories, with colors indicating their types: self-manipulations (yellow), self-protection behaviors (green), and manipulation of objects (blue). The three pie charts provide dataset-level statistics, including gender distribution, emotion labels of \textit{DEEMO-MER}, and three types of NFBL distributions of \textit{DEEMO-NFBL}. Best viewed digitally in color and zoomed in.}
    \label{fig:Histogram}
    \vspace{-1.5em}
\end{figure}

\begin{figure*}[t]
    \centering
    \includegraphics[width=0.9\linewidth]{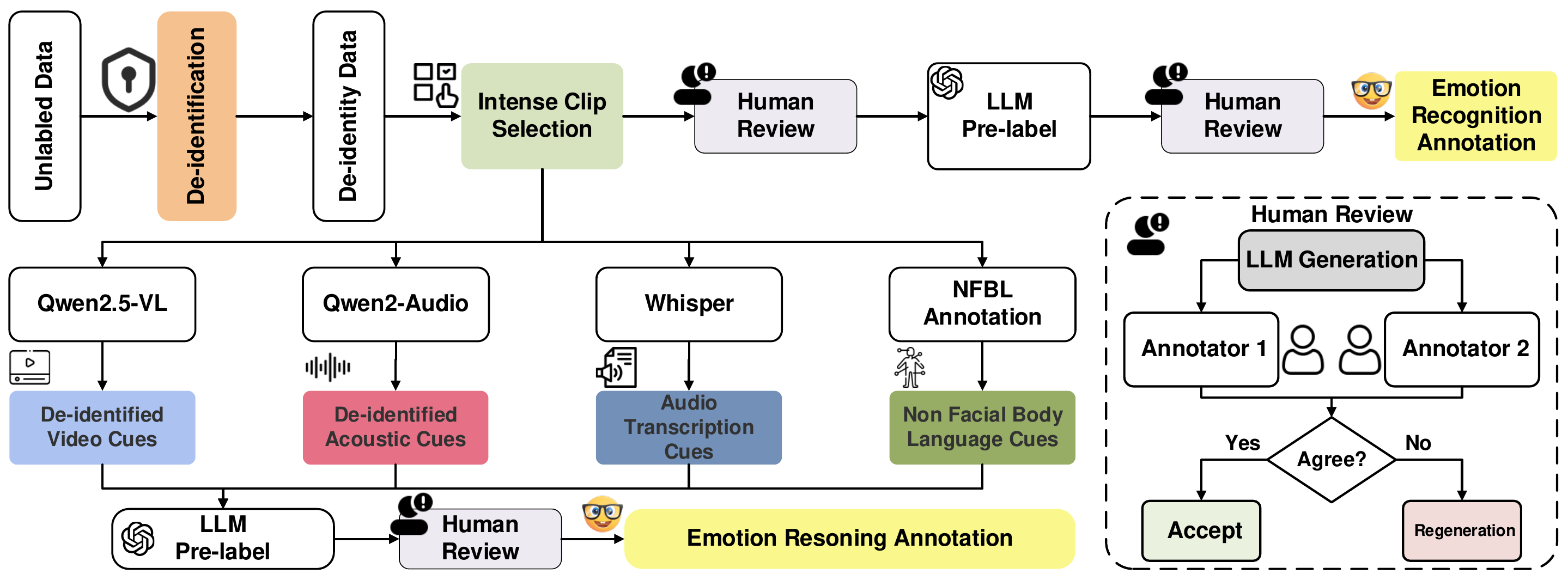}
    \caption{Overview of the annotation workflow for the \textit{DEEMO-MER}. The process consists of three stages: 1) data pre-labeling via multimodal de-identified cues (video, audio, transcription, and NFBL); 2) emotion recognition annotation through LLM-based pre-labeling followed by human review; 3) multimodal emotion reasoning annotation with LLM and human review. All LLM-generated annotations are cross-reviewed by two trained annotators to ensure labeling quality and consistency.
}
    \label{fig:pipeline}
\end{figure*}

\subsection{\textbf{DEEMO-NFBL}}
\subsubsection*{\textbf{DEEMO-NFBL Annotation.}}
After data collection and data de-identification, each video sequence is annotated with many NFBLs. NFBL has been shown to be an important cue for understanding subtle human emotions in affective computing~\cite{noroozi2018survey}. According to research on human behavior~\cite{navarro2008every}, NFBL includes self-manipulations (e.g., scratching the nose, touching the ear), manipulation of objects (e.g., playing with rings, pens, or paper), and self-protective behaviors (e.g., rubbing the eyes and folding arms). In total, 37 NFBL categories are defined, as illustrated in Fig.~\ref{fig:Histogram}. We annotate NFBL from the collected athlete interview videos and organize the resulting annotations into a subset of the dataset, which we refer to as \textit{DEEMO-NFBL}. The annotation process consists of two stages: First, the first group of trained annotators manually label NFBL instances in the videos, recording the start and end timestamps along with the NFBL type at the clip level. Then, a second group of trained annotators reviews and, if necessary, revises the annotations to ensure labeling accuracy and consistency.

\subsubsection*{\textbf{DEEMO-NFBL Statistics.}}
Table~\ref{tab:DEEMO-NFBL} summarizes the statistics of the \textit{DEEMO-NFBL} subset. It contains 500 long videos with a total of 24,722 annotated NFBL instances. The average video resolution is 1658×932 with a frame rate of 26.18 FPS. Each video lasts approximately 7 minutes on average, totaling 19.99 hours of de-identified videos. In addition, the \textit{DEEMO-NFBL} subset demonstrates cultural and demographic diversity. It includes 207 subjects from 75 different nationalities. Moreover, the gender distribution is relatively balanced, comprising 81 female and 126 male participants, which helps mitigate bias in downstream model training.

\begin{table}[]
    \centering
    \small
    \caption{Properties of the proposed \textit{DEEMO-NFBL} dataset.}
    \label{tab:DEEMO-NFBL}
    \vspace{-1em}
    \begin{tabular}{lr}
    \toprule
    \textbf{Properties}&\textbf{\textit{DEEMO-NFBL}}\\\hline
    Number of long videos&500\\
    Number of annotated NFBL&24722\\
    Resolution avg.&1658 $\times$ 932\\
    Frame rate avg.&26.18\\
    Subjects&207 (Female/Male 81/126)\\
    Nationality&75\\
    Total duration&19.99 (h)\\
    Duration avg.&416.05 (s)\\
    \bottomrule
    \end{tabular}
    \vspace{-2em}
\end{table}

\subsection{\textbf{DEEMO-MER}}
\subsubsection*{\textbf{De-identity Multimodal Emotion Recognition Annotation}}
Considering the nature of scenario post-game interviews with athletes, their emotional expressions may change throughout the video. Therefore, instead of assigning a single global emotion label to the entire video, we adopt a semi-automated annotation strategy that combines LLMs with human review to label emotionally intense clips more precisely. As shown in Fig.~\ref{fig:pipeline}, the annotation workflow of emotion recognition is as follows: 1) Given the strong text-understanding capabilities of LLMs, we first transcribe each video using a speech-to-text tool (e.g., Whisper~\cite{radford2023robust}), resulting in an audio transcription denoted as $T^{tr}$; 2) We then feed $T^{tr}$ into an LLM (e.g., GPT-4o), prompting it to identify one or more clips where the athlete exhibits strong emotional responses based on both the interview questions and the athlete’s replies. The prompt of LLM for intense emotion clip selection is present in supplementary material. Each selected clip is cross-reviewed by two human annotators. Specifically, two annotators cross-review each clip to ensure it reflects an intense emotional segment, as initially identified by GPT-4o. Intense clip is represented as a tuple $\left[t_s, t_e, T^{tr}_{[t_s, t_e]}, NFBL_{[t_s, t_e]}, A_{[t_s, t_e]}^{de}, V_{[t_s, t_e]}^{de}\right]$, where $t_s$ and $t_e$ denote the start and end timestamps, and $T^{tr}_{[t_s, t_e]}$, $A_{[t_s, t_e]}^{de}$, $V_{[t_s, t_e]}^{de}$ is the corresponding transcription, de-identified audio, de-identified video; 3) For each selected segment, the LLM assigns an initial emotional label (positive or negative). Two human annotators cross-review each labeled clip to verify the accuracy of the emotion classification. The resulting emotion-labeled clips form a subset of the \textit{DEEMO} dataset, which we name \textit{DEEMO-MER}.

\begin{table}[]
    \centering
    \small
    \caption{Properties of the proposed \textit{DEEMO-MER} dataset.}
    \vspace{-1em}
    \label{tab:DEEMO2K}
    \begin{tabular}{lr}
    \toprule
    \textbf{Properties}&\textbf{\textit{DEEMO-MER}}\\\hline
    Number of clips&2,060\\
    Number of positive samples&1,236\\
    Number of negative samples&824\\
    Frame rate avg.&26.18\\
    Total duration&7.12 (h)\\
    Duration avg.&11.29 (s)\\
    \bottomrule
    \end{tabular}
    \vspace{-1.5em}
\end{table}

\subsubsection*{\textbf{De-identity Multimodal Emotion Reasoning Annotation}}
Our goal is to enable the model to predict not only \textit{what} emotions are being expressed but also to infer \textit{why} these emotions occur. To this end, we further explore the use of the \textit{DEEMO-MER} dataset for a new task: De-identity Multimodal Emotion Reasoning. The pipeline of \textit{DEEMO-MER} annotation is shown in Fig.~\ref{fig:pipeline} and Alg. ~\ref{alg:deemo2k}.

\begin{algorithm}[H]
\footnotesize
\caption{Annotation Pipeline for \textit{DEEMO-MER}}
\label{alg:deemo2k}
\begin{algorithmic}[1]
\Require De-identified video $V^{\text{de}}$, de-identified audio $A^{\text{de}}$, Non Facial Body languages $NFBL$
\Ensure Multi-cues emotion labeled and reasoning  $Y_{\text{multi}}$

\State \textbf{Step 1: Transcription and Emotional Clip Extraction}
\State $T^{tr} \gets$ Whisper($A^{\text{de}}$)
\State $\mathcal{S} \gets$ GPT-4o.SelectIntenseEmotionalClips($T^{tr}$) 
\State $\mathcal{S} \gets$ HumanReview($\mathcal{S}$)

\ForAll{$[t_s, t_e] \in \mathcal{S}$}
    \State $V^{de}_{[t_s, t_e]}, A^{de}_{[t_s, t_e]}, T^{tr}_{[t_s, t_e]} \gets \text{Slice}(V^{\text{de}}, A^{\text{de}}, T^{tr}, [t_s, t_e])$
    \State \textbf{Step 2: Emotion Labeling}
    \State $D_{\text{emotion}} \gets$ GPT-4o.Prelabel($T^{tr}_{[t_s, t_e]}$)
    \State $Y_{\text{emotion}} \gets$ HumanReview($D^{\text{emotion}}_{[t_s, t_e]}$)

    \State \textbf{Step 3: Multimodal Cues Extraction for Reasoning}
    \State $D^{\text{visual}}_{[t_s, t_e]} \gets$Qwen2.5-VL.Extract($V_{[t_s, t_e]}^{\text{de}}$)
    \State $D^{\text{acoustic}}_{[t_s, t_e]} \gets$ Qwen2-Audio.Extract($A_{[t_s, t_e]}^{\text{de}}$)
    \State $NFBL_{[t_s, t_e]} \gets$ AnnotateNFBL($V_{[t_s, t_e]}^{\text{de}}$)

    \State \textbf{Step 4: Data Aggregation}
    \State $D^{\text{multi}}_{[t_s, t_e]} \gets$ GPT-4o.Aggregate($D^{\text{visual}}_{[t_s, t_e]}, D^{\text{acoustic}}_{[t_s, t_e]},
    T^{tr}_{[t_s, t_e]}, NFBL_{[t_s, t_e]}$)
    \State $Y^{\text{multi}}_{[t_s, t_e]} \gets$ HumanReview($D^{\text{multi}}_{[t_s, t_e]}$)
\EndFor
\end{algorithmic}
\end{algorithm}
\vspace{-1.5em}

Specifically, we adopt the following approach. First, Qwen2.5-VL~\cite{bai2025Qwen2} is used to extract visual contextual description $D^{\text{visual}}_{[t_s, t_e]}$ from the de-identified video clip $V^{\text{de}}_{[t_s, t_e]}$, including information such as characters, scenes, and environment which are elements that contribute to interpreting emotional background. 
Next, Qwen2-Audio~\cite{chu2024Qwen2} is employed to process the corresponding de-identified audio clip $A^{\text{de}}_{[t_s, t_e]}$ and extract emotion-relevant acoustic description $D^{\text{acoustic}}_{[t_s, t_e]}$, such as pitch and speech rate. Finally, we utilize \texttt{gpt-4o-2024-08-06}~\cite{achiam2023gpt} to aggregate the visual cues $D^{\text{visual}}_{[t_s, t_e]}$, acoustic cues $D^{\text{acoustic}}_{[t_s, t_e]}$, transcription text $T^{tr}_{[t_s, t_e]}$, and NFBL annotations $NFBL_{[t_s, t_e]}$ into a de-identity multimodal emotion reasoning description $D^{multi}_{[t_s, t_e]}$. To ensure the quality and reliability of the annotations, we introduce a human review step at multiple stages of the pipeline. Specifically, all emotional clips selected by the LLM and the emotion labels assigned to each clip are reviewed by two trained annotators. The annotators are instructed to verify whether the emotional clips are accurately selected and whether the assigned labels (positive or negative) are consistent with the multimodal emotion reasoning description $Y^{multi}_{[t_s, t_e]}$. The full annotation process for the \textit{DEEMO-MER} dataset, including both de-identified multimodal emotion recognition and reasoning preparation, is summarized in Algorithm~\ref{alg:deemo2k}. The prompt of MLLM (e.g, Qwen2.5-VL, Qwen2-audio, and GPT-4o) for visual, acoustic cues generation and multi-cues aggregation is present in the supplementary material. 

\textbf{DEEMO-MER Statistics.}
Table~\ref{tab:DEEMO2K} summarizes the statistics of the \textit{DEEMO-MER}. It contains 2,060 de-identified clips, including 1,236 labeled as positive and 824 as negative. In addition, 2060 emotion reasoning instructions are included. An example of emotion reasoning instruction is present in Figure~\ref{fig:framework}. 

\section{DEEMO-LLaMA}
\begin{figure*}[t]
    \centering
    \includegraphics[width=0.95\linewidth]{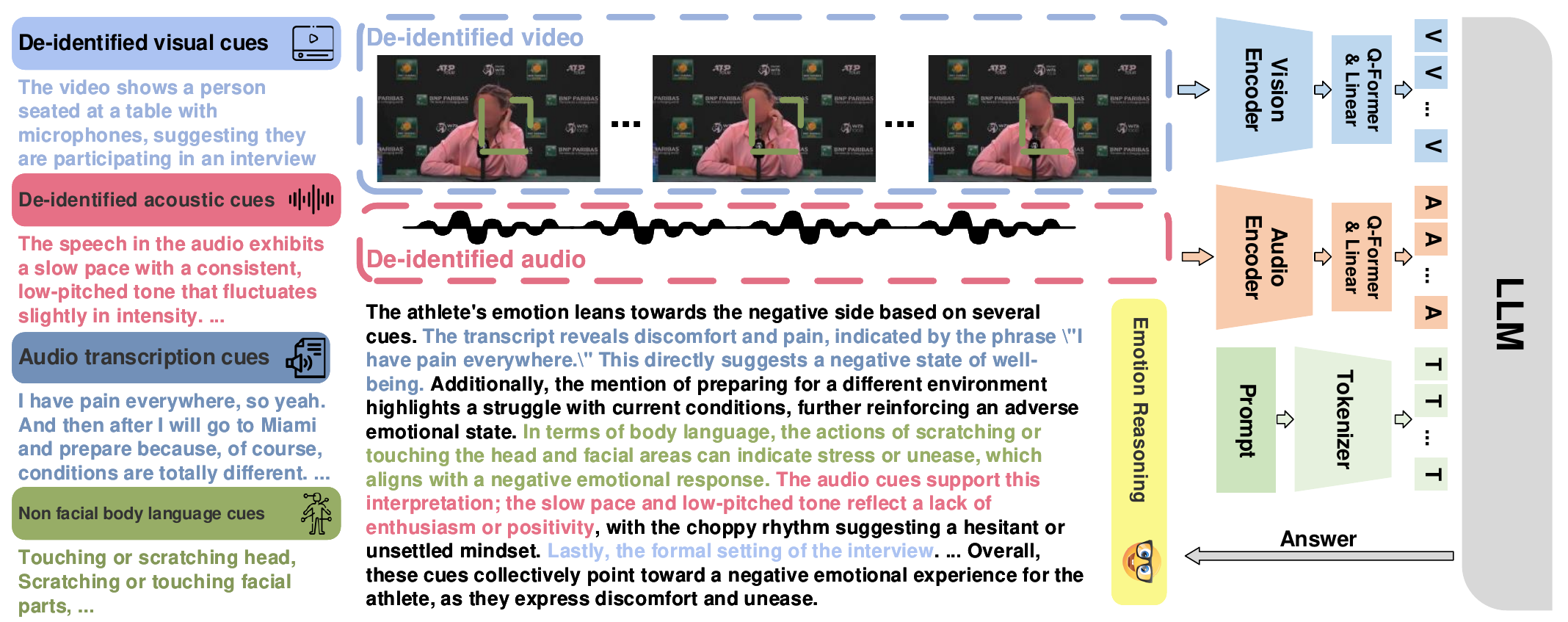}
    \vspace{-1em}
    \caption{Overview of the proposed DEEMO-LLaMA framework. The model takes de-identified video and audio as input. Each modality is processed by a dedicated encoder followed by a Q-former and a linear projection layer, and the resulting embeddings are integrated into a unified prompt for input to an LLM. In the emotion reasoning answer, different colors are used to highlight how DEEMO-LLaMA leverages various modality-specific cues (\textcolor{video}{visual cues}, \textcolor{audio}{acoustic cues}, \textcolor{trans}{audio transcription cues} and \textcolor{nfbl}{non facial body language cues}) to infer emotional states.}
    \label{fig:framework}
    \vspace{-1em}
\end{figure*}

To support the tasks introduced in \textit{DEEMO}, namely de-identity multimodal emotion recognition and de-identity multimodal emotion reasoning,
we propose an MLLM model called DEEMO-LLaMA, which accepts \textit{de-identified video, de-identified audio, and text} as inputs. The architecture of DEEMO-LLaMA is shown in Fig.~\ref{fig:framework}. 

\subsection{Architecture}

Given a de-identified video $V^{\text{de}}_{[t_s, t_e]}$ and de-identified audio $A^{\text{de}}_{[t_s, t_e]}$, we begin by extracting visual and audio features using pre-trained encoders. 
We uniformly sample $T$ frames $\{v_1^{\text{de}}, v_2^{\text{de}}, \dots, v_N^{\text{de}}\}, v_i \in \mathbb{R}^{H \times W \times 3}$ from the video. Each frame $v_i^{\text{de}}$ is processed using a frozen ViT-G/14 from EVA-CLIP~\cite{fang2023eva}. Then, we get $N$ tokens of dimension $d_v$ for each frame, resulting in $F_{\text{video}} \in \mathbb{R}^{TN \times d_v}$. Visual features $F_{\text{video}}$ are then passed into a pre-trained video Q-Former from BLIP-2~\cite{li2023blip}. 
The Q-Former compresses these into $K$ learnable query embeddings of dimension $d_q$, denoted as $\mathbf{V}_{q} \in \mathbb{R}^{K_v \times d_q}$. The Q-Former is used to compress relevant task-aware information from the dense visual features into learnable queries. 
For the audio modality, we utilize the pre-trained ImageBind~\cite{girdhar2023imagebind} as the audio encoder. Given a de-identified audio segment $A^{\text{de}}_{[t_s, t_e]}$, we uniformly sample $M$ short audio clips of 2 seconds $A^{\text{de}}_{[t_s, t_e]} = \{a_1, a_2, \dots, a_M\}, a_i \in \mathbb{R}^{F \times T_a}$, where $a_i$ denotes the spectrogram of the $i$-th audio clip, computed using 128 mel-spectrogram bins ($F = 128$) over $T_a$ time frames. 
We choose ImageBind~\cite{girdhar2023imagebind} as our audio encoder due to its strong generalization ability across modalities (such as audio-text).  Similar to video processing, we use an audio Q-Former to obtain audio embeddings $\mathbf{A}_{q} \in \mathbb{R}^{K_a \times d_q}$, where $K_a$ is the number of output queries and $d_q$ is the query embedding dimension. To integrate audio and visual embeddings with textual prompt input, we utilize a set of trainable linear projection layers that map video and audio embeddings into the LLM's input embedding space. $\mathbf{V} = \sigma_{\text{visual}}(\mathbf{V}_{q})$ and $\mathbf{A} = \sigma_{\text{audio}}(\mathbf{A}_{q})$ where $\sigma_{\text{visual}}$ and $\sigma_{\text{acoustic}}$ are modality-specific linear mappings. To format the input for emotion recognition and reasoning in a unified prompt style, we follow an instruction-based template that guides the LLM to attend to different modalities through structured tokens. Given visual tokens $\mathbf{V} \in \mathbb{R}^{K_v \times d_{\text{llm}}}$, audio tokens $\mathbf{A} \in \mathbb{R}^{K_a \times d_{\text{llm}}}$ and transcription $T^{tr}_{[t_s, t_e]}$, we construct the input sequence $X$ as:
\begin{equation}
\small
\label{eq:prompt}
\begin{aligned}
    X = [
        &\langle \texttt{Audio} \rangle \, \mathbf{A} \, \langle /\texttt{Audio} \rangle, \langle \texttt{Video} \rangle \, \mathbf{V} \, \langle /\texttt{Video} \rangle, \\
        &\langle \texttt{Transcription} \rangle \, T^{tr}_{[t_s, t_e]} \, \langle /\texttt{Transcription} \rangle,\\
        &\texttt{<User>}\, \text{Prompt} \, \texttt{</User>}]
\end{aligned}
\end{equation}
This multimodal input sequence $X$ is then processed by the LLM using its internal cross-attention layers, allowing it to capture emotional correlations across modalities and generate outputs for tasks, such as emotion recognition and reasoning explanations.

\subsection{Training}
\subsubsection*{\textbf{Pre-Training}}
We first conduct large-scale pre-training using multimodal datasets. Specifically, we leverage WebVid-2M~\cite{bain2021frozen}, a diverse collection of video-text pairs, and CC3M-595K~\cite{liu2023visual}, a image-caption data derived from CC3M~\cite{zhu2023minigpt}. These datasets provide rich visual semantics aligned with natural language descriptions. During this stage, the visual encoder and visual Q-Former are trained to encode visual content such that the frozen LLM can generate the corresponding captions. The task objective is formulated as a video-to-text generation task, where the model is encouraged to translate visual information into coherent natural language. Although the datasets may not fully capture the emotional or contextual depth of the video, this pre-training phase equips the model with a broad understanding of generic visual concepts.



\subsubsection*{\textbf{Multimodal Instruction Tuning}}
Following AffectGPT~\cite{lian2024affectgpt}, we further refined it through multimodal instruction tuning to enhance its capabilities in both emotion recognition and reasoning. This stage focuses on aligning the model's generative responses with di-identity emotion understanding. The instruction tuning is conducted on our proposed \textit{DEEMO-MER-training}. Each training sample is formatted as a multimodal prompt comprising audio tokens, visual tokens, and transcription, as illustrated in Eq.~\ref{eq:prompt}. The model is trained not only to predict the emotion label but also to generate a detailed explanation of the underlying emotional cause based on multimodal cues. The ID of the \textit{DEEMO-MER-training} videos is present in the supplemental material.
\section{Experiments}

\begin{table*}[t]
    \centering
    \footnotesize
    \caption{Ablation study on the impact of different input modalities.}
    \vspace{-1em}
    \label{tab:ablation}
    \begin{tabular}{cccccc}
    \toprule
         \textbf{Audio transcription}&\textbf{De-identified video}&\textbf{De-identified audio}&\textbf{NFBL}&\textbf{Accuracy (\%)}$\uparrow$&\textbf{F1-score (\%)}$\uparrow$\\
         \hline
         -&$\checkmark$&-&-&44.77&32.98\\
         -&-&$\checkmark$&-&53.19&50.12\\
         -&-&-&$\checkmark$&53.19&52.14\\
         \checkmark&-&-&-&66.67&63.13\\
         \checkmark&\checkmark&-&-&70.16&69.32\\
         \checkmark&\checkmark&\checkmark&-&74.49&74.45\\
         \checkmark&\checkmark&\checkmark&\checkmark&75.29&75.21\\
    \bottomrule
    \end{tabular}
    \vspace{-1em}
\end{table*}

\begin{table*}[t]
     \centering
     \footnotesize
     \caption{Comparison between DEEMO-LLaMA and other MLLMs. V, A, and T denote video, audio, and text modalities, respectively. F1-score and accuracy are measured in percentage (\%), while clue pverlap and label overlap are rated on a scale from 1 to 10.}
     \vspace{-1em}
     \label{tab:comparsion}
     \setlength\tabcolsep{1pt}
     \begin{tabular}{llccccccc}
     \toprule
          \multirow{2}{*}{\textbf{Method}}&\multirow{2}{*}{\textbf{Modality}}&\multirow{2}{*}{\textbf{LLM}}&\multirow{2}{*}{\textbf{Model size}}&\multicolumn{2}{c}{\textbf{Emotion Recognition}}&\multicolumn{3}{c}{\textbf{Emotion Reasoning}}  \\ 
          \cmidrule(lr){5-6} \cmidrule(lr){7-9} 
          &&&&\textbf{Accuracy$\uparrow$}
          &\textbf{F1-score$\uparrow$}&\textbf{Clue Overlap$\uparrow$}&\textbf{Label Overlap$\uparrow$}&\textbf{Avg$\uparrow$}\\
          \hline
          \multicolumn{9}{c}{\textit{General-purpose Video-based MLLMs}}\\
          VideoChatGPT~\cite{maazvideochatgpt}&V+T&Vicuna&7B&63.09&58.34&5.43&7.14&6.29\\
          Video-LLaVA~\cite{lin2023video}&V+T&Vicuna&7B&60.53&48.35&5.18&7.12&6.15\\
          Chat-UniVi~\cite{jin2024chat}&V+T&Vicuna&7B&58.24&43.50&5.18&6.60&5.89\\
          Movie-Chat~\cite{song2024moviechat}&V+T&Vicuna&7B&57.49&43.31&5.04&6.45&5.75\\
          \hline
          \multicolumn{9}{c}{\textit{General-purpose Video-Audio-based MLLMs}}\\
          Video-LLaMA~\cite{zhangvideollama}&V+A+T&Vicuna&7B&64.46&61.83&5.39&6.92&6.15\\
          PandaGPT~\cite{su2023pandagpt}&V+A+T&Vicuna&7B&61.49&52.61&5.45&6.77&6.11\\
          \hline
          \multicolumn{9}{c}{\textit{Emotional Video-Audio-based MLLMs}}\\
          Emotion-LLaMA~\cite{cheng2025emotion}&V+A+T&LLaMA 2&7B&63.41&61.58&5.41&7.22&6.31\\
          AffectGPT~\cite{lian2024affectgpt}&V+A+T&Vicuna&7B&64.74&62.09&5.12&6.66&5.89\\
          \textbf{DEEMO-LLaMA (Ours)}&V+A+T&Vicuna&7B&\textbf{74.49}&\textbf{74.45}&\textbf{6.20}&\textbf{7.66}&\textbf{6.93}\\
          \bottomrule
     \end{tabular}
     \vspace{-1em}
\end{table*}
\subsection{Metrics}
For the task of de-identity emotion recognition, the MLLM’s predictions are evaluated against the ground-truth emotion labels using standard classification metrics, including Top-1 accuracy and F1-score, defined as follows:
\begin{equation}
\begin{aligned}
    \text{Accuracy} & = \frac{TP + TN}{TP + TN + FP + FN}, \\
    \text{F1-score} & = \frac{2 \times \text{Precision} \times  \text{Recall}}{\text{Precision} + \text{Recall}}, \\
    \text{Precision} & = \frac{TP}{TP + FP}, \\
    \text{Recall} & = \frac{TP}{TP + FN}.
\end{aligned}
\end{equation}
For the task of de-identity multimodal emotion reasoning, we follow the GPT-based evaluation methodology proposed in Emotion-LLaMA~\cite{cheng2025emotion}. Specifically, we utilize \texttt{gpt-4o-mini-2024-07-18} to assess the model outputs by focusing on two key aspects: 1) \textbf{clue overlap} (1-10), the degree of overlap in emotion-related cues and the completeness of the cross-modal reasoning process, and 2) \textbf{label overlap} (1-10), the consistency of the summarized emotional states. The process of GPT-based evaluation of identity multimodal emotion reasoning is shown in supplementary material Algorithm~1. 

\subsection{Implementation Details}
Following Video-LLaMA~\cite{zhangvideollama}, we use the Vicuna-7B~\cite{vicuna2023} model as the language backbone. We uniformly sampled frames 8 per video, each resized to $224 \times 224$ and processed through a visual encoder, namely, ViT-G/14 from EVA-CLIP~\cite{fang2023eva}. The audio branch employs ImageBind~\cite{girdhar2023imagebind} for extracting de-identified acoustic embeddings. Then, Q-Formers~\cite{li2023blip} are used for both modalities, each producing a fixed number of learnable queries (32 for video, 8 for audio), which are further projected to the LLM token space via trainable linear layers. During multimodal instruction tuning, we freeze all encoder backbones and only fine-tune the video/audio Q-Formers and the linear projection heads. We use NVIDIA A100 GPU for training and inference. The batch size is 2.

\subsection{Ablation Study}
In the ablation study, we aim to answer two key questions: \textit{1) Does non-facial body language help for de-identity emotion understanding?} \textit{2) Is multimodal information necessary for de-identity emotion understandings?} To evaluate the impact of different cues, we progressively introduced various types of input cues during the inference, including 1) audio transcription, 2) de-identified video, 3) de-identified audio, and 4) NFBL. The NFBL information will be fed to MLLM in text form. Then, we compared their performance on the emotion recognition task by using \textit{DEEMO-MER-testing}. The ID of the \textit{DEEMO-MER-testing} clips is present in the supplemental material.

\subsubsection*{\textbf{Dose non-facial body language helpful?}}
As shown in Table~\ref{tab:ablation}, we first evaluate four de-identified cues independently: audio transcription, de-identified video, de-identified audio, and NFBL. Among all individual modalities, audio transcription performed the best, achieving 66.67\% accuracy. This suggests that the semantic content of the spoken responses retains strong emotional signals. NFBL slightly outperforms audio in terms of F1-score (52.14\% vs. 50.12\%), highlighting the effectiveness of NFBL in conveying affective states. This supports findings from psychology~\cite{noroozi2018survey} that emphasize the emotional expressiveness of NFBL. By contrast, only de-identified video has the lowest performance, indicating that visual context is less informative when facial expressions are masked.

\subsubsection*{\textbf{Is multimodal information necessary for de-identity emotion understandings?}}
As shown in Table~\ref{tab:ablation}, we take the unimodal setting using only audio transcription as the baseline. When additional modalities are introduced, performance consistently improves. For example, combining audio transcription, de-identified video, and de-identified audio achieves an accuracy of 74.49\% and an F1-score of 74.45\%. These results indicate that each modality contributes complementary information, and integrating them enhances the model's ability to recognize emotions. During multimodal instruction tuning, DEEMO-LLaMA implicitly learns to leverage NFBL cues. To further assess their impact, we additionally use explicit NFBL annotations in textual form as input. The observed performance gain confirms that that NFBL is an effective privacy-preserving cue for emotion understanding. However, due to the current lack of reliable NFBL detection models, we use audio transcription, de-identified video, and de-identified audio as the final input to our model, as shown in Figure~\ref{fig:framework}.

\subsection{Comparative Methods}

To evaluate the capability of recent MLLMs in the task of de-identity multimodal emotion understanding, we conduct a comprehensive comparison on the \textit{DEEMO-MER-testing} set. The evaluated MLLMs are categorized into three groups: 1) General-purpose Video-based MLLMs: including VideoChatGPT~\cite{maazvideochatgpt}, Video-LLaVA~\cite{lin2023video}, Chat-UniVi~\cite{jin2024chat}, and Movie-Chat~\cite{song2024moviechat}; 2) General-purpose Video-Audio-based MLLMs: including Video-LLaMA~\cite{zhangvideollama} and PandaGPT~\cite{su2023pandagpt}; Emotional Video-Audio-based MLLMs: including Emotion-LLaMA~\cite{cheng2025emotion} and AffectGPT~\cite{lian2024affectgpt}.To ensure fair comparisons, we use the 7B version for all models, follow their official implementations, and utilize their pre-trained weights. 

\begin{table*}[]
    \centering
    \footnotesize
    \caption{An example of de-identity multimodal emotion reasoning comparing the proposed DEEMO-LLaMA with other MLLMs. Incorrect reasoning is marked in red, and correct reasoning in green.}
    \begin{tabular}{l|p{12cm}}
    \toprule
         &\includegraphics[width=\linewidth]{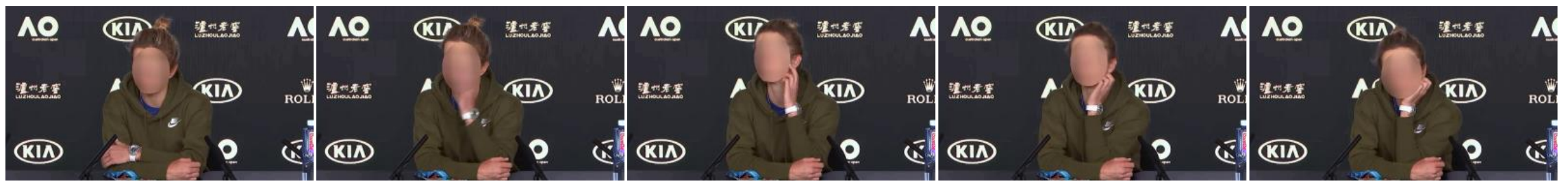}  \\
         Audio transcription&"She was striking the ball really good because I gave her this chance." \\\hline
         Ground Truth&"Based on the cues provided, the athlete's emotion appears to lean towards the negative. The transcription reveals a sense of blame or regret in the statement about her being given a chance, indicating that the athlete might not feel entirely satisfied or happy with the outcome. The body language shows illustrative hand gestures and the act of scratching or touching facial parts, which can be signs of nervousness or discomfort, potentially hinting at underlying frustration or disappointment. In addition, the audio cues suggest a calm and composed delivery, but the monotone quality of the speech, lacking variations in pitch or intensity, might reflect a sense of resignation rather than enthusiasm or positivity. Overall, the combination of these elements suggests an emotional state that is not particularly upbeat and leans towards a more negative sentiment regarding the situation."\\\hline
         Video-LLaMA~\cite{zhangvideollama}&"\textcolor{red}{The athlete's emotion is positive}. During the post-match interview, \textcolor{red}{she is smiling} and seems content with her performance. \textcolor{red}{She even thanks her coach for giving her the chance to strike the ball really well.} Therefore, her emotions are positive." \\
         \hline
         Emotion-LLaMA~\cite{su2023pandagpt}&"\textcolor{red}{the athlete's emotion is positive}, as \textcolor{red}{they express gratitude and acknowledgement towards their coach} for providing them with the opportunity to strike the ball well"\\\hline
         AffectGPT~\cite{zhangvideollama}&"In the video, an athlete wearing a green shirt can be seen sitting in a conference room with a microphone and a table full of various drinks and bottles. The athlete is interviewed by a journalist after a match. \textcolor{red}{The audio cues indicate that the athlete is speaking loudly and confidently,} with a positive tone. The video cues show the athlete sitting in a conference room, speaking into a microphone, and looking directly at the camera. \textcolor{red}{The non-facial body language cues include the athlete's posture, which is upright and confident}, and the fact that they are speaking loudly. \textcolor{red}{Overall, the video and audio suggest a positive and confident mood from the athlete.}"\\\hline
         \textbf{DEEMO-LLaMA (Ours)}&"Based on the provided cues, the athlete's emotion is \textcolor{correct}{more likely `negative.'} The transcription indicates a focus on what could have been done better, \textcolor{correct}{specifically emphasizing "gave her this chance," which suggests disappointment} or frustration with a performance. The non-facial body language, specifically \textcolor{correct}{touching or scratching the face, often indicates discomfort or anxiety}, which aligns with a negative emotional state. \textcolor{correct}{In the audio, the slow-paced, monotone delivery lacking in fluctuations could imply a deflated or depressed mood}, contrasting with the uplifting emotions associated with a positive experience. \textcolor{correct}{Collectively, these elements lead to the inference that the athlete is grappling with negative feelings about the situation.}"\\\bottomrule
    \end{tabular}
    \label{tab:case}
\end{table*}

\subsubsection*{\textbf{De-identity Mutlimoda Emotion Recognition.}} 
In the Table~\ref{tab:comparsion},  DEEMO-LLaMA achieves the best performance on the emotion recognition task, with an accuracy of 74.49\% and an F1-score of 74.45\%, significantly outperforming all other models across the three MLLM categories. Compared with general-purpose video-only MLLMs (e.g., VideoChatGPT, Video-LLaVA, Chat-UniVi, and Movie-Chat), models that incorporate audio inputs (e.g., Video-LLaMA, PandaGPT) perform notably better, which highlights the importance of acoustic information. However, even among emotional video-audio MLLMs, such as Emotion-LLaMA and AffectGPT, performance remains limited. This suggests that existing MLLMs struggle to generalize emotion understanding in a privacy-preserving, de-identified emotion understanding. In contrast, DEEMO-LLaMA effectively leverages all available identity-agnostic cues (transcription, acoustic, visual, and NFBL). The prompt for MLLMs of emotion recognition is present in the supplementary material.

\subsubsection*{\textbf{De-identity Mutlimoda Emotion Reasoning.}}
DEEMO-LLaMA achieves the best performance with a clue overlap of 6.20, a label overlap of 7.66, and an overall average of 6.93, as shown in Table~\ref{tab:comparsion}. These results suggest that DEEMO-LLaMA is not only capable of correctly identifying emotional state, but also excels at generating coherent and well-grounded reasoning based on multimodal cues (e.g., visual, acoustic, and NFBL). The prompt for MLLMs of emotion reasoning is present in the supplementary material.

\subsection{Qualitative Analysis}
To better understand the reasoning ability of different MLLMs under the de-identity setting, we provide a qualitative comparison in Table~\ref{tab:case}. 
Given the same de-identified video, audio, and transcription inputs, only DEEMO-LLaMA correctly infers a negative emotional state and supports its inference using diverse modality-specific evidence. 
Specifically, it captures the cues: 1) The athlete’s phrasing (“gave her this chance”) may imply regret or frustration; 2) NFBL (e.g., touching or scratching the face), which conveys anxiety; 3) Acoustic patterns like a low-pitched, monotonous delivery, which reflect a subdued emotional tone. 
In contrast, other MLLMs, such as VideoLLaMA, Emotion-LLaMA, and AffectGPT, incorrectly predict a positive emotion, relying primarily on superficial features such as word sentiment or posture while overlooking more subtle, affect-rich signals. This highlights a key advantage of DEEMO-LLaMA: its ability to reason over implicit, privacy-preserving cues and deliver well-grounded reasoning.

\section{Conclusion}
In this work, we introduce novel tasks called \textbf{De}-identity Multimodal \textbf{Emo}tion Recognition and Reasoning. To support this task, we construct the \textbf{\textit{DEEMO}} dataset, which includes two subsets: \textit{DEEMO-NFBL}, which contains 24,722 Non Facial Body Language (NFBL) annotations, and \textit{DEEMO-MER}, which includes 2,060 annotated videos with both emotion labels and reasoning instructions through LLM-human collaborative annotation. To the best of our knowledge, \textit{DEEMO} is the first dataset specifically designed for privacy-preserving emotion understanding for Multimodal Large Language Models (MLLMs). Building on this foundation, we propose DEEMO-LLaMA, a unified MLLM that integrates de-identified cues across video, audio, and text. Through extensive experiments, we demonstrate that DEEMO-LLaMA achieves state-of-the-art performance on both de-identified emotion recognition and emotion reasoning, outperforming existing MLLMs by a significant margin. Our results highlight the importance of leveraging implicit and privacy-preserving emotional signals for interpretable affective modeling. We hope that our dataset and framework can inspire future research in privacy-preserving emotion understanding and ethical human-centered AI.

\newpage
\newpage

\bibliographystyle{ACM-Reference-Format}
\bibliography{main}


\begin{thebibliography}{64}


\ifx \showCODEN    \undefined \def \showCODEN     #1{\unskip}     \fi
\ifx \showDOI      \undefined \def \showDOI       #1{#1}\fi
\ifx \showISBNx    \undefined \def \showISBNx     #1{\unskip}     \fi
\ifx \showISBNxiii \undefined \def \showISBNxiii  #1{\unskip}     \fi
\ifx \showISSN     \undefined \def \showISSN      #1{\unskip}     \fi
\ifx \showLCCN     \undefined \def \showLCCN      #1{\unskip}     \fi
\ifx \shownote     \undefined \def \shownote      #1{#1}          \fi
\ifx \showarticletitle \undefined \def \showarticletitle #1{#1}   \fi
\ifx \showURL      \undefined \def \showURL       {\relax}        \fi
\providecommand\bibfield[2]{#2}
\providecommand\bibinfo[2]{#2}
\providecommand\natexlab[1]{#1}
\providecommand\showeprint[2][]{arXiv:#2}

\bibitem[Achiam et~al\mbox{.}(2023)]%
        {achiam2023gpt}
\bibfield{author}{\bibinfo{person}{Josh Achiam}, \bibinfo{person}{Steven Adler}, \bibinfo{person}{Sandhini Agarwal}, \bibinfo{person}{Lama Ahmad}, \bibinfo{person}{Ilge Akkaya}, \bibinfo{person}{Florencia~Leoni Aleman}, \bibinfo{person}{Diogo Almeida}, \bibinfo{person}{Janko Altenschmidt}, \bibinfo{person}{Sam Altman}, \bibinfo{person}{Shyamal Anadkat}, {et~al\mbox{.}}} \bibinfo{year}{2023}\natexlab{}.
\newblock \showarticletitle{Gpt-4 technical report}.
\newblock \bibinfo{journal}{\emph{arXiv preprint arXiv:2303.08774}} (\bibinfo{year}{2023}).
\newblock


\bibitem[Bai et~al\mbox{.}(2025)]%
        {bai2025Qwen2}
\bibfield{author}{\bibinfo{person}{Shuai Bai}, \bibinfo{person}{Keqin Chen}, \bibinfo{person}{Xuejing Liu}, \bibinfo{person}{Jialin Wang}, \bibinfo{person}{Wenbin Ge}, \bibinfo{person}{Sibo Song}, \bibinfo{person}{Kai Dang}, \bibinfo{person}{Peng Wang}, \bibinfo{person}{Shijie Wang}, \bibinfo{person}{Jun Tang}, {et~al\mbox{.}}} \bibinfo{year}{2025}\natexlab{}.
\newblock \showarticletitle{Qwen2. 5-vl technical report}.
\newblock \bibinfo{journal}{\emph{arXiv preprint arXiv:2502.13923}} (\bibinfo{year}{2025}).
\newblock


\bibitem[Bain et~al\mbox{.}(2021)]%
        {bain2021frozen}
\bibfield{author}{\bibinfo{person}{Max Bain}, \bibinfo{person}{Arsha Nagrani}, \bibinfo{person}{G{\"u}l Varol}, {and} \bibinfo{person}{Andrew Zisserman}.} \bibinfo{year}{2021}\natexlab{}.
\newblock \showarticletitle{Frozen in time: A joint video and image encoder for end-to-end retrieval}. In \bibinfo{booktitle}{\emph{IEEE/CVF International Conference on Computer Vision}}. \bibinfo{pages}{1728--1738}.
\newblock


\bibitem[Baveye et~al\mbox{.}(2015)]%
        {baveye2015liris}
\bibfield{author}{\bibinfo{person}{Yoann Baveye}, \bibinfo{person}{Emmanuel Dellandrea}, \bibinfo{person}{Christel Chamaret}, {and} \bibinfo{person}{Liming Chen}.} \bibinfo{year}{2015}\natexlab{}.
\newblock \showarticletitle{LIRIS-ACCEDE: A video database for affective content analysis}.
\newblock \bibinfo{journal}{\emph{IEEE Transactions on Affective Computing}} \bibinfo{volume}{6}, \bibinfo{number}{1} (\bibinfo{year}{2015}), \bibinfo{pages}{43--55}.
\newblock


\bibitem[Busso et~al\mbox{.}(2008)]%
        {busso2008iemocap}
\bibfield{author}{\bibinfo{person}{Carlos Busso}, \bibinfo{person}{Murtaza Bulut}, \bibinfo{person}{Chi-Chun Lee}, \bibinfo{person}{Abe Kazemzadeh}, \bibinfo{person}{Emily Mower}, \bibinfo{person}{Samuel Kim}, \bibinfo{person}{Jeannette~N Chang}, \bibinfo{person}{Sungbok Lee}, {and} \bibinfo{person}{Shrikanth~S Narayanan}.} \bibinfo{year}{2008}\natexlab{}.
\newblock \showarticletitle{IEMOCAP: Interactive emotional dyadic motion capture database}.
\newblock \bibinfo{journal}{\emph{Language Resources and Evaluation}}  \bibinfo{volume}{42} (\bibinfo{year}{2008}), \bibinfo{pages}{335--359}.
\newblock


\bibitem[Canal et~al\mbox{.}(2022)]%
        {canal2022survey}
\bibfield{author}{\bibinfo{person}{Felipe~Zago Canal}, \bibinfo{person}{Tobias~Rossi M{\"u}ller}, \bibinfo{person}{Jhennifer~Cristine Matias}, \bibinfo{person}{Gustavo~Gino Scotton}, \bibinfo{person}{Antonio~Reis de Sa~Junior}, \bibinfo{person}{Eliane Pozzebon}, {and} \bibinfo{person}{Antonio~Carlos Sobieranski}.} \bibinfo{year}{2022}\natexlab{}.
\newblock \showarticletitle{A survey on facial emotion recognition techniques: A state-of-the-art literature review}.
\newblock \bibinfo{journal}{\emph{Information Sciences}}  \bibinfo{volume}{582} (\bibinfo{year}{2022}), \bibinfo{pages}{593--617}.
\newblock


\bibitem[Chen et~al\mbox{.}(2023)]%
        {chen2023smg}
\bibfield{author}{\bibinfo{person}{Haoyu Chen}, \bibinfo{person}{Henglin Shi}, \bibinfo{person}{Xin Liu}, \bibinfo{person}{Xiaobai Li}, {and} \bibinfo{person}{Guoying Zhao}.} \bibinfo{year}{2023}\natexlab{}.
\newblock \showarticletitle{SMG: A micro-gesture dataset towards spontaneous body gestures for emotional stress state analysis}.
\newblock \bibinfo{journal}{\emph{International Journal of Computer Vision}} \bibinfo{volume}{131}, \bibinfo{number}{6} (\bibinfo{year}{2023}), \bibinfo{pages}{1346--1366}.
\newblock


\bibitem[Cheng et~al\mbox{.}(2025)]%
        {cheng2025emotion}
\bibfield{author}{\bibinfo{person}{Zebang Cheng}, \bibinfo{person}{Zhi-Qi Cheng}, \bibinfo{person}{Jun-Yan He}, \bibinfo{person}{Kai Wang}, \bibinfo{person}{Yuxiang Lin}, \bibinfo{person}{Zheng Lian}, \bibinfo{person}{Xiaojiang Peng}, {and} \bibinfo{person}{Alexander Hauptmann}.} \bibinfo{year}{2025}\natexlab{}.
\newblock \showarticletitle{Emotion-llama: Multimodal emotion recognition and reasoning with instruction tuning}.
\newblock \bibinfo{journal}{\emph{Advances in Neural Information Processing Systems}}  \bibinfo{volume}{37} (\bibinfo{year}{2025}), \bibinfo{pages}{110805--110853}.
\newblock


\bibitem[Chiang et~al\mbox{.}(2023)]%
        {vicuna2023}
\bibfield{author}{\bibinfo{person}{Wei-Lin Chiang}, \bibinfo{person}{Zhuohan Li}, \bibinfo{person}{Zi Lin}, \bibinfo{person}{Ying Sheng}, \bibinfo{person}{Zhanghao Wu}, \bibinfo{person}{Hao Zhang}, \bibinfo{person}{Lianmin Zheng}, \bibinfo{person}{Siyuan Zhuang}, \bibinfo{person}{Yonghao Zhuang}, \bibinfo{person}{Joseph~E. Gonzalez}, \bibinfo{person}{Ion Stoica}, {and} \bibinfo{person}{Eric~P. Xing}.} \bibinfo{year}{2023}\natexlab{}.
\newblock \bibinfo{title}{Vicuna: An Open-Source Chatbot Impressing GPT-4 with 90\%* ChatGPT Quality}.
\newblock
\newblock
\urldef\tempurl%
\url{https://lmsys.org/blog/2023-03-30-vicuna/}
\showURL{%
\tempurl}


\bibitem[Chu et~al\mbox{.}(2024)]%
        {chu2024Qwen2}
\bibfield{author}{\bibinfo{person}{Yunfei Chu}, \bibinfo{person}{Jin Xu}, \bibinfo{person}{Qian Yang}, \bibinfo{person}{Haojie Wei}, \bibinfo{person}{Xipin Wei}, \bibinfo{person}{Zhifang Guo}, \bibinfo{person}{Yichong Leng}, \bibinfo{person}{Yuanjun Lv}, \bibinfo{person}{Jinzheng He}, \bibinfo{person}{Junyang Lin}, {et~al\mbox{.}}} \bibinfo{year}{2024}\natexlab{}.
\newblock \showarticletitle{Qwen2-audio technical report}.
\newblock \bibinfo{journal}{\emph{arXiv preprint arXiv:2407.10759}} (\bibinfo{year}{2024}).
\newblock


\bibitem[Dhall et~al\mbox{.}(2012)]%
        {dhall2012collecting}
\bibfield{author}{\bibinfo{person}{Abhinav Dhall}, \bibinfo{person}{Roland Goecke}, \bibinfo{person}{Simon Lucey}, {and} \bibinfo{person}{Tom Gedeon}.} \bibinfo{year}{2012}\natexlab{}.
\newblock \showarticletitle{Collecting Large, Richly Annotated Facial-Expression Databases from Movies}.
\newblock \bibinfo{journal}{\emph{IEEE MultiMedia}} \bibinfo{volume}{19}, \bibinfo{number}{3} (\bibinfo{year}{2012}), \bibinfo{pages}{34--41}.
\newblock
\urldef\tempurl%
\url{https://doi.org/10.1109/MMUL.2012.26}
\showDOI{\tempurl}


\bibitem[Dhall et~al\mbox{.}(2015)]%
        {dhall2015video}
\bibfield{author}{\bibinfo{person}{Abhinav Dhall}, \bibinfo{person}{OV Ramana~Murthy}, \bibinfo{person}{Roland Goecke}, \bibinfo{person}{Jyoti Joshi}, {and} \bibinfo{person}{Tom Gedeon}.} \bibinfo{year}{2015}\natexlab{}.
\newblock \showarticletitle{Video and image based emotion recognition challenges in the wild: Emotiw 2015}. In \bibinfo{booktitle}{\emph{ACM on International Conference on Multimodal Interaction}}. \bibinfo{pages}{423--426}.
\newblock


\bibitem[Douglas-Cowie et~al\mbox{.}(2007)]%
        {douglas2007humaine}
\bibfield{author}{\bibinfo{person}{Ellen Douglas-Cowie}, \bibinfo{person}{Roddy Cowie}, \bibinfo{person}{Ian Sneddon}, \bibinfo{person}{Cate Cox}, \bibinfo{person}{Orla Lowry}, \bibinfo{person}{Margaret Mcrorie}, \bibinfo{person}{Jean-Claude Martin}, \bibinfo{person}{Laurence Devillers}, \bibinfo{person}{Sarkis Abrilian}, \bibinfo{person}{Anton Batliner}, {et~al\mbox{.}}} \bibinfo{year}{2007}\natexlab{}.
\newblock \showarticletitle{The HUMAINE database: Addressing the collection and annotation of naturalistic and induced emotional data}. In \bibinfo{booktitle}{\emph{International Conference on Affective Computing and Intelligent Interaction}}. Springer, \bibinfo{pages}{488--500}.
\newblock


\bibitem[El~Ayadi et~al\mbox{.}(2011)]%
        {el2011survey}
\bibfield{author}{\bibinfo{person}{Moataz El~Ayadi}, \bibinfo{person}{Mohamed~S Kamel}, {and} \bibinfo{person}{Fakhri Karray}.} \bibinfo{year}{2011}\natexlab{}.
\newblock \showarticletitle{Survey on speech emotion recognition: Features, classification schemes, and databases}.
\newblock \bibinfo{journal}{\emph{Pattern recognition}} \bibinfo{volume}{44}, \bibinfo{number}{3} (\bibinfo{year}{2011}), \bibinfo{pages}{572--587}.
\newblock


\bibitem[Ezzameli and Mahersia(2023)]%
        {ezzameli2023emotion}
\bibfield{author}{\bibinfo{person}{K Ezzameli} {and} \bibinfo{person}{H Mahersia}.} \bibinfo{year}{2023}\natexlab{}.
\newblock \showarticletitle{Emotion recognition from unimodal to multimodal analysis: A review}.
\newblock \bibinfo{journal}{\emph{Information Fusion}} (\bibinfo{year}{2023}), \bibinfo{pages}{101847}.
\newblock


\bibitem[Fang et~al\mbox{.}(2023)]%
        {fang2023eva}
\bibfield{author}{\bibinfo{person}{Yuxin Fang}, \bibinfo{person}{Wen Wang}, \bibinfo{person}{Binhui Xie}, \bibinfo{person}{Quan Sun}, \bibinfo{person}{Ledell Wu}, \bibinfo{person}{Xinggang Wang}, \bibinfo{person}{Tiejun Huang}, \bibinfo{person}{Xinlong Wang}, {and} \bibinfo{person}{Yue Cao}.} \bibinfo{year}{2023}\natexlab{}.
\newblock \showarticletitle{Eva: Exploring the limits of masked visual representation learning at scale}. In \bibinfo{booktitle}{\emph{IEEE/CVF Conference on Computer Vision and Pattern Recognition}}. \bibinfo{pages}{19358--19369}.
\newblock


\bibitem[Fourati and Pelachaud(2014)]%
        {fourati2014emilya}
\bibfield{author}{\bibinfo{person}{Nesrine Fourati} {and} \bibinfo{person}{Catherine Pelachaud}.} \bibinfo{year}{2014}\natexlab{}.
\newblock \showarticletitle{Emilya: Emotional body expression in daily actions database.}. In \bibinfo{booktitle}{\emph{LREC}}. \bibinfo{pages}{3486--3493}.
\newblock


\bibitem[Gao et~al\mbox{.}(2024)]%
        {gao2024identity}
\bibfield{author}{\bibinfo{person}{Rong Gao}, \bibinfo{person}{Xin Liu}, \bibinfo{person}{Bohao Xing}, \bibinfo{person}{Zitong Yu}, \bibinfo{person}{Bjorn~W Schuller}, {and} \bibinfo{person}{Heikki K{\"a}lvi{\"a}inen}.} \bibinfo{year}{2024}\natexlab{}.
\newblock \showarticletitle{Identity-free artificial emotional intelligence via micro-gesture understanding}.
\newblock \bibinfo{journal}{\emph{arXiv preprint arXiv:2405.13206}} (\bibinfo{year}{2024}).
\newblock


\bibitem[Girdhar et~al\mbox{.}(2023)]%
        {girdhar2023imagebind}
\bibfield{author}{\bibinfo{person}{Rohit Girdhar}, \bibinfo{person}{Alaaeldin El-Nouby}, \bibinfo{person}{Zhuang Liu}, \bibinfo{person}{Mannat Singh}, \bibinfo{person}{Kalyan~Vasudev Alwala}, \bibinfo{person}{Armand Joulin}, {and} \bibinfo{person}{Ishan Misra}.} \bibinfo{year}{2023}\natexlab{}.
\newblock \showarticletitle{Imagebind: One embedding space to bind them all}. In \bibinfo{booktitle}{\emph{Proceedings of the IEEE/CVF conference on computer vision and pattern recognition}}. \bibinfo{pages}{15180--15190}.
\newblock


\bibitem[Grimm et~al\mbox{.}(2008)]%
        {grimm2008vera}
\bibfield{author}{\bibinfo{person}{Michael Grimm}, \bibinfo{person}{Kristian Kroschel}, {and} \bibinfo{person}{Shrikanth Narayanan}.} \bibinfo{year}{2008}\natexlab{}.
\newblock \showarticletitle{The Vera am Mittag German audio-visual emotional speech database}. In \bibinfo{booktitle}{\emph{IEEE International Conference on Multimedia and Expo}}. IEEE, \bibinfo{pages}{865--868}.
\newblock


\bibitem[Guo et~al\mbox{.}(2024)]%
        {guo2024benchmarking}
\bibfield{author}{\bibinfo{person}{Dan Guo}, \bibinfo{person}{Kun Li}, \bibinfo{person}{Bin Hu}, \bibinfo{person}{Yan Zhang}, {and} \bibinfo{person}{Meng Wang}.} \bibinfo{year}{2024}\natexlab{}.
\newblock \showarticletitle{Benchmarking micro-action recognition: Dataset, methods, and applications}.
\newblock \bibinfo{journal}{\emph{IEEE Transactions on Circuits and Systems for Video Technology}} \bibinfo{volume}{34}, \bibinfo{number}{7} (\bibinfo{year}{2024}), \bibinfo{pages}{6238--6252}.
\newblock


\bibitem[Hakak et~al\mbox{.}(2017)]%
        {hakak2017emotion}
\bibfield{author}{\bibinfo{person}{Nida~Manzoor Hakak}, \bibinfo{person}{Mohsin Mohd}, \bibinfo{person}{Mahira Kirmani}, {and} \bibinfo{person}{Mudasir Mohd}.} \bibinfo{year}{2017}\natexlab{}.
\newblock \showarticletitle{Emotion analysis: A survey}. In \bibinfo{booktitle}{\emph{2017 international conference on computer, communications and electronics (COMPTELIX)}}. IEEE, \bibinfo{pages}{397--402}.
\newblock


\bibitem[Hashem et~al\mbox{.}(2023)]%
        {hashem2023speech}
\bibfield{author}{\bibinfo{person}{Ahlam Hashem}, \bibinfo{person}{Muhammad Arif}, {and} \bibinfo{person}{Manal Alghamdi}.} \bibinfo{year}{2023}\natexlab{}.
\newblock \showarticletitle{Speech emotion recognition approaches: A systematic review}.
\newblock \bibinfo{journal}{\emph{Speech Communication}}  \bibinfo{volume}{154} (\bibinfo{year}{2023}), \bibinfo{pages}{102974}.
\newblock


\bibitem[Hsu et~al\mbox{.}(2017)]%
        {hsu2017automatic}
\bibfield{author}{\bibinfo{person}{Yu-Liang Hsu}, \bibinfo{person}{Jeen-Shing Wang}, \bibinfo{person}{Wei-Chun Chiang}, {and} \bibinfo{person}{Chien-Han Hung}.} \bibinfo{year}{2017}\natexlab{}.
\newblock \showarticletitle{Automatic ECG-based emotion recognition in music listening}.
\newblock \bibinfo{journal}{\emph{IEEE Transactions on Affective Computing}} \bibinfo{volume}{11}, \bibinfo{number}{1} (\bibinfo{year}{2017}), \bibinfo{pages}{85--99}.
\newblock


\bibitem[Jin et~al\mbox{.}(2024)]%
        {jin2024chat}
\bibfield{author}{\bibinfo{person}{Peng Jin}, \bibinfo{person}{Ryuichi Takanobu}, \bibinfo{person}{Wancai Zhang}, \bibinfo{person}{Xiaochun Cao}, {and} \bibinfo{person}{Li Yuan}.} \bibinfo{year}{2024}\natexlab{}.
\newblock \showarticletitle{Chat-univi: Unified visual representation empowers large language models with image and video understanding}. In \bibinfo{booktitle}{\emph{Proceedings of the IEEE/CVF Conference on Computer Vision and Pattern Recognition}}. \bibinfo{pages}{13700--13710}.
\newblock


\bibitem[Koelstra et~al\mbox{.}(2011)]%
        {koelstra2011deap}
\bibfield{author}{\bibinfo{person}{Sander Koelstra}, \bibinfo{person}{Christian Muhl}, \bibinfo{person}{Mohammad Soleymani}, \bibinfo{person}{Jong-Seok Lee}, \bibinfo{person}{Ashkan Yazdani}, \bibinfo{person}{Touradj Ebrahimi}, \bibinfo{person}{Thierry Pun}, \bibinfo{person}{Anton Nijholt}, {and} \bibinfo{person}{Ioannis Patras}.} \bibinfo{year}{2011}\natexlab{}.
\newblock \showarticletitle{Deap: A database for emotion analysis; using physiological signals}.
\newblock \bibinfo{journal}{\emph{IEEE Transactions on Affective Computing}} \bibinfo{volume}{3}, \bibinfo{number}{1} (\bibinfo{year}{2011}), \bibinfo{pages}{18--31}.
\newblock


\bibitem[Kossaifi et~al\mbox{.}(2019)]%
        {kossaifi2019sewa}
\bibfield{author}{\bibinfo{person}{Jean Kossaifi}, \bibinfo{person}{Robert Walecki}, \bibinfo{person}{Yannis Panagakis}, \bibinfo{person}{Jie Shen}, \bibinfo{person}{Maximilian Schmitt}, \bibinfo{person}{Fabien Ringeval}, \bibinfo{person}{Jing Han}, \bibinfo{person}{Vedhas Pandit}, \bibinfo{person}{Antoine Toisoul}, \bibinfo{person}{Bj{\"o}rn Schuller}, {et~al\mbox{.}}} \bibinfo{year}{2019}\natexlab{}.
\newblock \showarticletitle{Sewa db: A rich database for audio-visual emotion and sentiment research in the wild}.
\newblock \bibinfo{journal}{\emph{IEEE Transactions on Pattern Analysis and Machine Intelligence}} \bibinfo{volume}{43}, \bibinfo{number}{3} (\bibinfo{year}{2019}), \bibinfo{pages}{1022--1040}.
\newblock


\bibitem[Krumhuber et~al\mbox{.}(2023)]%
        {krumhuber2023role}
\bibfield{author}{\bibinfo{person}{Eva~G Krumhuber}, \bibinfo{person}{Lina~I Skora}, \bibinfo{person}{Harold~CH Hill}, {and} \bibinfo{person}{Karen Lander}.} \bibinfo{year}{2023}\natexlab{}.
\newblock \showarticletitle{The role of facial movements in emotion recognition}.
\newblock \bibinfo{journal}{\emph{Nature Reviews Psychology}} \bibinfo{volume}{2}, \bibinfo{number}{5} (\bibinfo{year}{2023}), \bibinfo{pages}{283--296}.
\newblock


\bibitem[Li et~al\mbox{.}(2024)]%
        {li2024enhancing}
\bibfield{author}{\bibinfo{person}{Deng Li}, \bibinfo{person}{Bohao Xing}, {and} \bibinfo{person}{Xin Liu}.} \bibinfo{year}{2024}\natexlab{}.
\newblock \showarticletitle{Enhancing micro gesture recognition for emotion understanding via context-aware visual-text contrastive learning}.
\newblock \bibinfo{journal}{\emph{IEEE Signal Processing Letters}} (\bibinfo{year}{2024}).
\newblock


\bibitem[Li et~al\mbox{.}(2023)]%
        {li2023blip}
\bibfield{author}{\bibinfo{person}{Junnan Li}, \bibinfo{person}{Dongxu Li}, \bibinfo{person}{Silvio Savarese}, {and} \bibinfo{person}{Steven Hoi}.} \bibinfo{year}{2023}\natexlab{}.
\newblock \showarticletitle{Blip-2: Bootstrapping language-image pre-training with frozen image encoders and large language models}. In \bibinfo{booktitle}{\emph{International conference on machine learning}}. PMLR, \bibinfo{pages}{19730--19742}.
\newblock


\bibitem[Li and Deng(2020)]%
        {li2020deep}
\bibfield{author}{\bibinfo{person}{Shan Li} {and} \bibinfo{person}{Weihong Deng}.} \bibinfo{year}{2020}\natexlab{}.
\newblock \showarticletitle{Deep facial expression recognition: A survey}.
\newblock \bibinfo{journal}{\emph{IEEE Transactions on Affective Computing}} \bibinfo{volume}{13}, \bibinfo{number}{3} (\bibinfo{year}{2020}), \bibinfo{pages}{1195--1215}.
\newblock


\bibitem[Li et~al\mbox{.}(2022)]%
        {li2022eeg}
\bibfield{author}{\bibinfo{person}{Xiang Li}, \bibinfo{person}{Yazhou Zhang}, \bibinfo{person}{Prayag Tiwari}, \bibinfo{person}{Dawei Song}, \bibinfo{person}{Bin Hu}, \bibinfo{person}{Meihong Yang}, \bibinfo{person}{Zhigang Zhao}, \bibinfo{person}{Neeraj Kumar}, {and} \bibinfo{person}{Pekka Marttinen}.} \bibinfo{year}{2022}\natexlab{}.
\newblock \showarticletitle{EEG based emotion recognition: A tutorial and review}.
\newblock \bibinfo{journal}{\emph{Comput. Surveys}} \bibinfo{volume}{55}, \bibinfo{number}{4} (\bibinfo{year}{2022}), \bibinfo{pages}{1--57}.
\newblock


\bibitem[Lian et~al\mbox{.}(2024)]%
        {lian2024affectgpt}
\bibfield{author}{\bibinfo{person}{Zheng Lian}, \bibinfo{person}{Haiyang Sun}, \bibinfo{person}{Licai Sun}, \bibinfo{person}{Jiangyan Yi}, \bibinfo{person}{Bin Liu}, {and} \bibinfo{person}{Jianhua Tao}.} \bibinfo{year}{2024}\natexlab{}.
\newblock \showarticletitle{AffectGPT: Dataset and framework for explainable multimodal emotion recognition}.
\newblock \bibinfo{journal}{\emph{arXiv preprint arXiv:2407.07653}} (\bibinfo{year}{2024}).
\newblock


\bibitem[Lian et~al\mbox{.}(2023)]%
        {lian2023explainable}
\bibfield{author}{\bibinfo{person}{Zheng Lian}, \bibinfo{person}{Licai Sun}, \bibinfo{person}{Mingyu Xu}, \bibinfo{person}{Haiyang Sun}, \bibinfo{person}{Ke Xu}, \bibinfo{person}{Zhuofan Wen}, \bibinfo{person}{Shun Chen}, \bibinfo{person}{Bin Liu}, {and} \bibinfo{person}{Jianhua Tao}.} \bibinfo{year}{2023}\natexlab{}.
\newblock \showarticletitle{Explainable multimodal emotion reasoning}.
\newblock \bibinfo{journal}{\emph{CoRR}} (\bibinfo{year}{2023}).
\newblock


\bibitem[Lin et~al\mbox{.}(2023)]%
        {lin2023video}
\bibfield{author}{\bibinfo{person}{Bin Lin}, \bibinfo{person}{Yang Ye}, \bibinfo{person}{Bin Zhu}, \bibinfo{person}{Jiaxi Cui}, \bibinfo{person}{Munan Ning}, \bibinfo{person}{Peng Jin}, {and} \bibinfo{person}{Li Yuan}.} \bibinfo{year}{2023}\natexlab{}.
\newblock \showarticletitle{Video-llava: Learning united visual representation by alignment before projection}.
\newblock \bibinfo{journal}{\emph{arXiv preprint arXiv:2311.10122}} (\bibinfo{year}{2023}).
\newblock


\bibitem[Liu et~al\mbox{.}(2023)]%
        {liu2023visual}
\bibfield{author}{\bibinfo{person}{Haotian Liu}, \bibinfo{person}{Chunyuan Li}, \bibinfo{person}{Qingyang Wu}, {and} \bibinfo{person}{Yong~Jae Lee}.} \bibinfo{year}{2023}\natexlab{}.
\newblock \showarticletitle{Visual instruction tuning}.
\newblock \bibinfo{journal}{\emph{Advances in Neural Information Processing systems}}  \bibinfo{volume}{36} (\bibinfo{year}{2023}), \bibinfo{pages}{34892--34916}.
\newblock


\bibitem[Liu et~al\mbox{.}(2016)]%
        {liu2016retracted}
\bibfield{author}{\bibinfo{person}{Mingyang Liu}, \bibinfo{person}{Di Fan}, \bibinfo{person}{Xiaohan Zhang}, {and} \bibinfo{person}{Xiaopeng Gong}.} \bibinfo{year}{2016}\natexlab{}.
\newblock \showarticletitle{Retracted: Human emotion recognition based on galvanic skin response signal feature selection and svm}. In \bibinfo{booktitle}{\emph{International Conference on Smart City and Systems Engineering}}. IEEE, \bibinfo{pages}{157--160}.
\newblock


\bibitem[Liu et~al\mbox{.}(2021a)]%
        {liu2021comparing}
\bibfield{author}{\bibinfo{person}{Wei Liu}, \bibinfo{person}{Jie-Lin Qiu}, \bibinfo{person}{Wei-Long Zheng}, {and} \bibinfo{person}{Bao-Liang Lu}.} \bibinfo{year}{2021}\natexlab{a}.
\newblock \showarticletitle{Comparing recognition performance and robustness of multimodal deep learning models for multimodal emotion recognition}.
\newblock \bibinfo{journal}{\emph{IEEE Transactions on Cognitive and Developmental Systems}} \bibinfo{volume}{14}, \bibinfo{number}{2} (\bibinfo{year}{2021}), \bibinfo{pages}{715--729}.
\newblock


\bibitem[Liu et~al\mbox{.}(2021b)]%
        {liu2021imigue}
\bibfield{author}{\bibinfo{person}{Xin Liu}, \bibinfo{person}{Henglin Shi}, \bibinfo{person}{Haoyu Chen}, \bibinfo{person}{Zitong Yu}, \bibinfo{person}{Xiaobai Li}, {and} \bibinfo{person}{Guoying Zhao}.} \bibinfo{year}{2021}\natexlab{b}.
\newblock \showarticletitle{iMiGUE: An identity-free video dataset for micro-gesture understanding and emotion analysis}. In \bibinfo{booktitle}{\emph{IEEE/CVF Conference on Computer Vision and Pattern Recognition}}. \bibinfo{pages}{10631--10642}.
\newblock


\bibitem[Maaz et~al\mbox{.}(2024)]%
        {maazvideochatgpt}
\bibfield{author}{\bibinfo{person}{Muhammad Maaz}, \bibinfo{person}{Hanoona Rasheed}, \bibinfo{person}{Salman Khan}, {and} \bibinfo{person}{Fahad Khan}.} \bibinfo{year}{2024}\natexlab{}.
\newblock \showarticletitle{Video-{C}hat{GPT}: Towards Detailed Video Understanding via Large Vision and Language Models}. In \bibinfo{booktitle}{\emph{62nd Annual Meeting of the Association for Computational Linguistics}}. \bibinfo{publisher}{Association for Computational Linguistics}, \bibinfo{address}{Bangkok, Thailand}, \bibinfo{pages}{12585--12602}.
\newblock


\bibitem[McAdams(1984)]%
        {mcadams1984spectral}
\bibfield{author}{\bibinfo{person}{Stephen~Edward McAdams}.} \bibinfo{year}{1984}\natexlab{}.
\newblock \bibinfo{booktitle}{\emph{Spectral fusion, spectral parsing and the formation of auditory images}}.
\newblock \bibinfo{publisher}{Stanford university}.
\newblock


\bibitem[McDuff et~al\mbox{.}(2013)]%
        {mcduff2013affectiva}
\bibfield{author}{\bibinfo{person}{Daniel McDuff}, \bibinfo{person}{Rana Kaliouby}, \bibinfo{person}{Thibaud Senechal}, \bibinfo{person}{May Amr}, \bibinfo{person}{Jeffrey Cohn}, {and} \bibinfo{person}{Rosalind Picard}.} \bibinfo{year}{2013}\natexlab{}.
\newblock \showarticletitle{Affectiva-mit facial expression dataset (am-fed): Naturalistic and spontaneous facial expressions collected}. In \bibinfo{booktitle}{\emph{IEEE Conference on Computer Vision and Pattern Recognition Workshops}}. \bibinfo{pages}{881--888}.
\newblock


\bibitem[Meden et~al\mbox{.}(2021)]%
        {meden2021privacy}
\bibfield{author}{\bibinfo{person}{Bla{\v{z}} Meden}, \bibinfo{person}{Peter Rot}, \bibinfo{person}{Philipp Terh{\"o}rst}, \bibinfo{person}{Naser Damer}, \bibinfo{person}{Arjan Kuijper}, \bibinfo{person}{Walter~J Scheirer}, \bibinfo{person}{Arun Ross}, \bibinfo{person}{Peter Peer}, {and} \bibinfo{person}{Vitomir {\v{S}}truc}.} \bibinfo{year}{2021}\natexlab{}.
\newblock \showarticletitle{Privacy--enhancing face biometrics: A comprehensive survey}.
\newblock \bibinfo{journal}{\emph{IEEE Transactions on Information Forensics and Security}}  \bibinfo{volume}{16} (\bibinfo{year}{2021}), \bibinfo{pages}{4147--4183}.
\newblock


\bibitem[Middya et~al\mbox{.}(2022)]%
        {middya2022deep}
\bibfield{author}{\bibinfo{person}{Asif~Iqbal Middya}, \bibinfo{person}{Baibhav Nag}, {and} \bibinfo{person}{Sarbani Roy}.} \bibinfo{year}{2022}\natexlab{}.
\newblock \showarticletitle{Deep learning based multimodal emotion recognition using model-level fusion of audio--visual modalities}.
\newblock \bibinfo{journal}{\emph{Knowledge-based systems}}  \bibinfo{volume}{244} (\bibinfo{year}{2022}), \bibinfo{pages}{108580}.
\newblock


\bibitem[Morency et~al\mbox{.}(2011)]%
        {morency2011towards}
\bibfield{author}{\bibinfo{person}{Louis-Philippe Morency}, \bibinfo{person}{Rada Mihalcea}, {and} \bibinfo{person}{Payal Doshi}.} \bibinfo{year}{2011}\natexlab{}.
\newblock \showarticletitle{Towards multimodal sentiment analysis: Harvesting opinions from the web}. In \bibinfo{booktitle}{\emph{International Conference on Multimodal Interfaces}}. \bibinfo{pages}{169--176}.
\newblock


\bibitem[Nandwani and Verma(2021)]%
        {nandwani2021review}
\bibfield{author}{\bibinfo{person}{Pansy Nandwani} {and} \bibinfo{person}{Rupali Verma}.} \bibinfo{year}{2021}\natexlab{}.
\newblock \showarticletitle{A review on sentiment analysis and emotion detection from text}.
\newblock \bibinfo{journal}{\emph{Social network analysis and mining}} \bibinfo{volume}{11}, \bibinfo{number}{1} (\bibinfo{year}{2021}), \bibinfo{pages}{81}.
\newblock


\bibitem[Navarro and Karlins(2008)]%
        {navarro2008every}
\bibfield{author}{\bibinfo{person}{Julia Navarro} {and} \bibinfo{person}{Marvin Karlins}.} \bibinfo{year}{2008}\natexlab{}.
\newblock \bibinfo{booktitle}{\emph{What every body is saying}}.
\newblock \bibinfo{publisher}{HarperCollins Publishers New York, NY, USA:}.
\newblock


\bibitem[Naveed et~al\mbox{.}(2023)]%
        {naveed2023comprehensive}
\bibfield{author}{\bibinfo{person}{Humza Naveed}, \bibinfo{person}{Asad~Ullah Khan}, \bibinfo{person}{Shi Qiu}, \bibinfo{person}{Muhammad Saqib}, \bibinfo{person}{Saeed Anwar}, \bibinfo{person}{Muhammad Usman}, \bibinfo{person}{Naveed Akhtar}, \bibinfo{person}{Nick Barnes}, {and} \bibinfo{person}{Ajmal Mian}.} \bibinfo{year}{2023}\natexlab{}.
\newblock \showarticletitle{A comprehensive overview of large language models}.
\newblock \bibinfo{journal}{\emph{arXiv preprint arXiv:2307.06435}} (\bibinfo{year}{2023}).
\newblock


\bibitem[Noroozi et~al\mbox{.}(2018)]%
        {noroozi2018survey}
\bibfield{author}{\bibinfo{person}{Fatemeh Noroozi}, \bibinfo{person}{Ciprian~Adrian Corneanu}, \bibinfo{person}{Dorota Kami{\'n}ska}, \bibinfo{person}{Tomasz Sapi{\'n}ski}, \bibinfo{person}{Sergio Escalera}, {and} \bibinfo{person}{Gholamreza Anbarjafari}.} \bibinfo{year}{2018}\natexlab{}.
\newblock \showarticletitle{Survey on emotional body gesture recognition}.
\newblock \bibinfo{journal}{\emph{IEEE transactions on affective computing}} \bibinfo{volume}{12}, \bibinfo{number}{2} (\bibinfo{year}{2018}), \bibinfo{pages}{505--523}.
\newblock


\bibitem[Radford et~al\mbox{.}(2023)]%
        {radford2023robust}
\bibfield{author}{\bibinfo{person}{Alec Radford}, \bibinfo{person}{Jong~Wook Kim}, \bibinfo{person}{Tao Xu}, \bibinfo{person}{Greg Brockman}, \bibinfo{person}{Christine McLeavey}, {and} \bibinfo{person}{Ilya Sutskever}.} \bibinfo{year}{2023}\natexlab{}.
\newblock \showarticletitle{Robust speech recognition via large-scale weak supervision}. In \bibinfo{booktitle}{\emph{International Conference on Machine Learning}}. PMLR, \bibinfo{pages}{28492--28518}.
\newblock


\bibitem[Rahdari et~al\mbox{.}(2019)]%
        {rahdari2019multimodal}
\bibfield{author}{\bibinfo{person}{Farhad Rahdari}, \bibinfo{person}{Esmat Rashedi}, {and} \bibinfo{person}{Mahdi Eftekhari}.} \bibinfo{year}{2019}\natexlab{}.
\newblock \showarticletitle{A multimodal emotion recognition system using facial landmark analysis}.
\newblock \bibinfo{journal}{\emph{Iranian Journal of Science and Technology, Transactions of Electrical Engineering}}  \bibinfo{volume}{43} (\bibinfo{year}{2019}), \bibinfo{pages}{171--189}.
\newblock


\bibitem[Rathi and Tripathy(2024)]%
        {rathi2024analyzing}
\bibfield{author}{\bibinfo{person}{Tarun Rathi} {and} \bibinfo{person}{Manoj Tripathy}.} \bibinfo{year}{2024}\natexlab{}.
\newblock \showarticletitle{Analyzing the influence of different speech data corpora and speech features on speech emotion recognition: A review}.
\newblock \bibinfo{journal}{\emph{Speech Communication}} (\bibinfo{year}{2024}), \bibinfo{pages}{103102}.
\newblock


\bibitem[Regulation(2018)]%
        {regulation2018general}
\bibfield{author}{\bibinfo{person}{Protection Regulation}.} \bibinfo{year}{2018}\natexlab{}.
\newblock \showarticletitle{General data protection regulation}.
\newblock \bibinfo{journal}{\emph{Intouch}}  \bibinfo{volume}{25} (\bibinfo{year}{2018}), \bibinfo{pages}{1--5}.
\newblock


\bibitem[Sanderson and Paliwal(2004)]%
        {sanderson2004use}
\bibfield{author}{\bibinfo{person}{Conrad Sanderson} {and} \bibinfo{person}{Kuldip~K Paliwal}.} \bibinfo{year}{2004}\natexlab{}.
\newblock \showarticletitle{On the use of speech and face information for identity verification}.
\newblock  (\bibinfo{year}{2004}).
\newblock


\bibitem[Singh and Goel(2022)]%
        {singh2022systematic}
\bibfield{author}{\bibinfo{person}{Youddha~Beer Singh} {and} \bibinfo{person}{Shivani Goel}.} \bibinfo{year}{2022}\natexlab{}.
\newblock \showarticletitle{A systematic literature review of speech emotion recognition approaches}.
\newblock \bibinfo{journal}{\emph{Neurocomputing}}  \bibinfo{volume}{492} (\bibinfo{year}{2022}), \bibinfo{pages}{245--263}.
\newblock


\bibitem[Song et~al\mbox{.}(2024)]%
        {song2024moviechat}
\bibfield{author}{\bibinfo{person}{Enxin Song}, \bibinfo{person}{Wenhao Chai}, \bibinfo{person}{Guanhong Wang}, \bibinfo{person}{Yucheng Zhang}, \bibinfo{person}{Haoyang Zhou}, \bibinfo{person}{Feiyang Wu}, \bibinfo{person}{Haozhe Chi}, \bibinfo{person}{Xun Guo}, \bibinfo{person}{Tian Ye}, \bibinfo{person}{Yanting Zhang}, {et~al\mbox{.}}} \bibinfo{year}{2024}\natexlab{}.
\newblock \showarticletitle{Moviechat: From dense token to sparse memory for long video understanding}. In \bibinfo{booktitle}{\emph{IEEE/CVF Conference on Computer Vision and Pattern Recognition}}. \bibinfo{pages}{18221--18232}.
\newblock


\bibitem[Su et~al\mbox{.}(2023)]%
        {su2023pandagpt}
\bibfield{author}{\bibinfo{person}{Yixuan Su}, \bibinfo{person}{Tian Lan}, \bibinfo{person}{Huayang Li}, \bibinfo{person}{Jialu Xu}, \bibinfo{person}{Yan Wang}, {and} \bibinfo{person}{Deng Cai}.} \bibinfo{year}{2023}\natexlab{}.
\newblock \showarticletitle{Pandagpt: One model to instruction-follow them all}.
\newblock \bibinfo{journal}{\emph{arXiv preprint arXiv:2305.16355}} (\bibinfo{year}{2023}).
\newblock


\bibitem[Tomashenko et~al\mbox{.}(2024)]%
        {tomashenko2024voiceprivacy}
\bibfield{author}{\bibinfo{person}{Natalia Tomashenko}, \bibinfo{person}{Xiaoxiao Miao}, \bibinfo{person}{Pierre Champion}, \bibinfo{person}{Sarina Meyer}, \bibinfo{person}{Xin Wang}, \bibinfo{person}{Emmanuel Vincent}, \bibinfo{person}{Michele Panariello}, \bibinfo{person}{Nicholas Evans}, \bibinfo{person}{Junichi Yamagishi}, {and} \bibinfo{person}{Massimiliano Todisco}.} \bibinfo{year}{2024}\natexlab{}.
\newblock \showarticletitle{The {VoicePrivacy} 2024 Challenge Evaluation Plan}.
\newblock  (\bibinfo{year}{2024}).
\newblock
\showeprint[arxiv]{2404.02677}~[eess.AS]


\bibitem[Wei et~al\mbox{.}(2024)]%
        {wei2024learning}
\bibfield{author}{\bibinfo{person}{Jie Wei}, \bibinfo{person}{Guanyu Hu}, \bibinfo{person}{Xinyu Yang}, \bibinfo{person}{Anh~Tuan Luu}, {and} \bibinfo{person}{Yizhuo Dong}.} \bibinfo{year}{2024}\natexlab{}.
\newblock \showarticletitle{Learning facial expression and body gesture visual information for video emotion recognition}.
\newblock \bibinfo{journal}{\emph{Expert Systems with Applications}}  \bibinfo{volume}{237} (\bibinfo{year}{2024}), \bibinfo{pages}{121419}.
\newblock


\bibitem[Xing et~al\mbox{.}(2024)]%
        {xing2024emo}
\bibfield{author}{\bibinfo{person}{Bohao Xing}, \bibinfo{person}{Zitong Yu}, \bibinfo{person}{Xin Liu}, \bibinfo{person}{Kaishen Yuan}, \bibinfo{person}{Qilang Ye}, \bibinfo{person}{Weicheng Xie}, \bibinfo{person}{Huanjing Yue}, \bibinfo{person}{Jingyu Yang}, {and} \bibinfo{person}{Heikki K{\"a}lvi{\"a}inen}.} \bibinfo{year}{2024}\natexlab{}.
\newblock \showarticletitle{Emo-llama: Enhancing facial emotion understanding with instruction tuning}.
\newblock \bibinfo{journal}{\emph{arXiv preprint arXiv:2408.11424}} (\bibinfo{year}{2024}).
\newblock


\bibitem[Xu et~al\mbox{.}(2020)]%
        {xu2020centerface}
\bibfield{author}{\bibinfo{person}{Yuanyuan Xu}, \bibinfo{person}{Wan Yan}, \bibinfo{person}{Genke Yang}, \bibinfo{person}{Jiliang Luo}, \bibinfo{person}{Tao Li}, {and} \bibinfo{person}{Jianan He}.} \bibinfo{year}{2020}\natexlab{}.
\newblock \showarticletitle{CenterFace: joint face detection and alignment using face as point}.
\newblock \bibinfo{journal}{\emph{Scientific Programming}} \bibinfo{volume}{2020}, \bibinfo{number}{1} (\bibinfo{year}{2020}), \bibinfo{pages}{7845384}.
\newblock


\bibitem[Zadeh et~al\mbox{.}(2018)]%
        {zadeh2018multimodal}
\bibfield{author}{\bibinfo{person}{AmirAli~Bagher Zadeh}, \bibinfo{person}{Paul~Pu Liang}, \bibinfo{person}{Soujanya Poria}, \bibinfo{person}{Erik Cambria}, {and} \bibinfo{person}{Louis-Philippe Morency}.} \bibinfo{year}{2018}\natexlab{}.
\newblock \showarticletitle{Multimodal language analysis in the wild: Cmu-mosei dataset and interpretable dynamic fusion graph}. In \bibinfo{booktitle}{\emph{Annual Meeting of the Association for Computational Linguistics}}. \bibinfo{pages}{2236--2246}.
\newblock


\bibitem[Zhang et~al\mbox{.}(2023)]%
        {zhangvideollama}
\bibfield{author}{\bibinfo{person}{Hang Zhang}, \bibinfo{person}{Xin Li}, {and} \bibinfo{person}{Lidong Bing}.} \bibinfo{year}{2023}\natexlab{}.
\newblock \showarticletitle{Video-{LL}a{MA}: An Instruction-tuned Audio-Visual Language Model for Video Understanding}. In \bibinfo{booktitle}{\emph{2023 Conference on Empirical Methods in Natural Language Processing: System Demonstrations}}. \bibinfo{publisher}{Association for Computational Linguistics}, \bibinfo{address}{Singapore}, \bibinfo{pages}{543--553}.
\newblock


\bibitem[Zhu et~al\mbox{.}(2023)]%
        {zhu2023minigpt}
\bibfield{author}{\bibinfo{person}{Deyao Zhu}, \bibinfo{person}{Jun Chen}, \bibinfo{person}{Xiaoqian Shen}, \bibinfo{person}{Xiang Li}, {and} \bibinfo{person}{Mohamed Elhoseiny}.} \bibinfo{year}{2023}\natexlab{}.
\newblock \showarticletitle{Minigpt-4: Enhancing vision-language understanding with advanced large language models}.
\newblock \bibinfo{journal}{\emph{arXiv preprint arXiv:2304.10592}} (\bibinfo{year}{2023}).
\newblock


\end{thebibliography}
\end{document}